\documentclass{article}

\PassOptionsToPackage{numbers, compress}{natbib}
 \usepackage[preprint]{neurips_2026}

\usepackage[utf8]{inputenc} % allow utf-8 input
\usepackage[T1]{fontenc}    % use 8-bit T1 fonts
\usepackage{hyperref}       % hyperlinks
\usepackage{url}            % simple URL typesetting
\usepackage{booktabs}       % professional-quality tables
\usepackage{amsfonts}       % blackboard math symbols
\usepackage{nicefrac}       % compact symbols for 1/2, etc.
\usepackage{microtype}      % microtypography
\usepackage{xcolor}         % colors

\usepackage{graphicx}
\newcommand{\eg}{\textit{e.g.}}
\usepackage{amsmath}
\usepackage{algorithm}
\usepackage{algpseudocode}
\usepackage{enumitem}
\usepackage{wrapfig}
\usepackage{multirow}
\usepackage[table]{xcolor}
\title{Rosetta: Composable Native Multimodal Pretraining}

\author{%
  {\bf Xiangyue Liu}\textsuperscript{1} \quad 
  {\bf Zijian Zhang}\textsuperscript{2} \quad 
  {\bf Miles Yang}\textsuperscript{2} \quad 
  {\bf Zhao Zhong}\textsuperscript{2} \\[1.5mm]
  {\bf Liefeng Bo}\textsuperscript{2} \quad 
  {\bf Ping Tan}\textsuperscript{1}\thanks{Corresponding author.} \\[2.5mm]
  \textsuperscript{1}HKUST \quad \textsuperscript{2}Tencent Hunyuan
  \vspace{-4mm} % 微微调紧，确保与下方 Abstract 保持最紧凑、体面的过渡
}

\begin{document}

\maketitle

\begin{abstract}
    Achieving true artificial general intelligence requires foundation models capable of integrating new modalities without forgetting prior knowledge. However, accommodating continuous generative objectives alongside discrete understanding tasks causes severe gradient conflicts. Existing architectures, including standard Mixture-of-Experts (MoE), are highly susceptible to representation overwriting. Even structurally partitioned paradigms like Mixture-of-Transformers (MoT) remain vulnerable to catastrophic forgetting, severely impeding multimodal scalability. In this work, we introduce Rosetta, a composable native multimodal pretraining framework designed for seamless and non-destructive modality expansion. Rosetta adopts a modular paradigm where core foundational knowledge is preserved within global shared experts, while modality-specific capabilities are distributed across plug-and-play experts. To guarantee non-destructive composition, we propose Momentum-Anchored Orthogonal Projection (MAOP). MAOP leverages the optimizer's momentum state as an implicit semantic anchor, selectively neutralizing conflicting gradient components from new modalities while preserving synergistic updates. To strictly isolate the architectural impact, we evaluate Rosetta against standard MoE and MoT baselines under strict active parameter parity. All models are trained from scratch within the Transfusion framework, using discrete next-token prediction for language and continuous visual diffusion. Extensive evaluations demonstrate that, while standard MoE and MoT architectures suffer catastrophic forgetting of previously acquired knowledge, Rosetta robustly preserves established language and visual understanding. Furthermore, it delivers superior image generation and unlocks cross-modal synergy, paving the way for truly composable and unified multimodal foundation models. To facilitate further multimodal research, we release our code and checkpoints to the community.
    Project page at {\hypersetup{pdfborder={0 0 0}}\href{https://rosetta-lmm.github.io/}{\color{magenta}\texttt{https://rosetta-lmm.github.io/}}}.
    % Project page at {\hypersetup{pdfborder={0 0 0}}\href{https://rosetta-lmm.github.io/}{\color{blue}\texttt{https://rosetta-lmm.github.io}}}.
    
    % natively composable multimodal foundation models.
    % composable, unified, and extensible multimodal foundation models.
\end{abstract}

\section{Introduction}
\label{sec:intro}

\begin{figure}[t!] 
    \centering
    \vspace{-2em} 
    \includegraphics[width=1.0\textwidth]{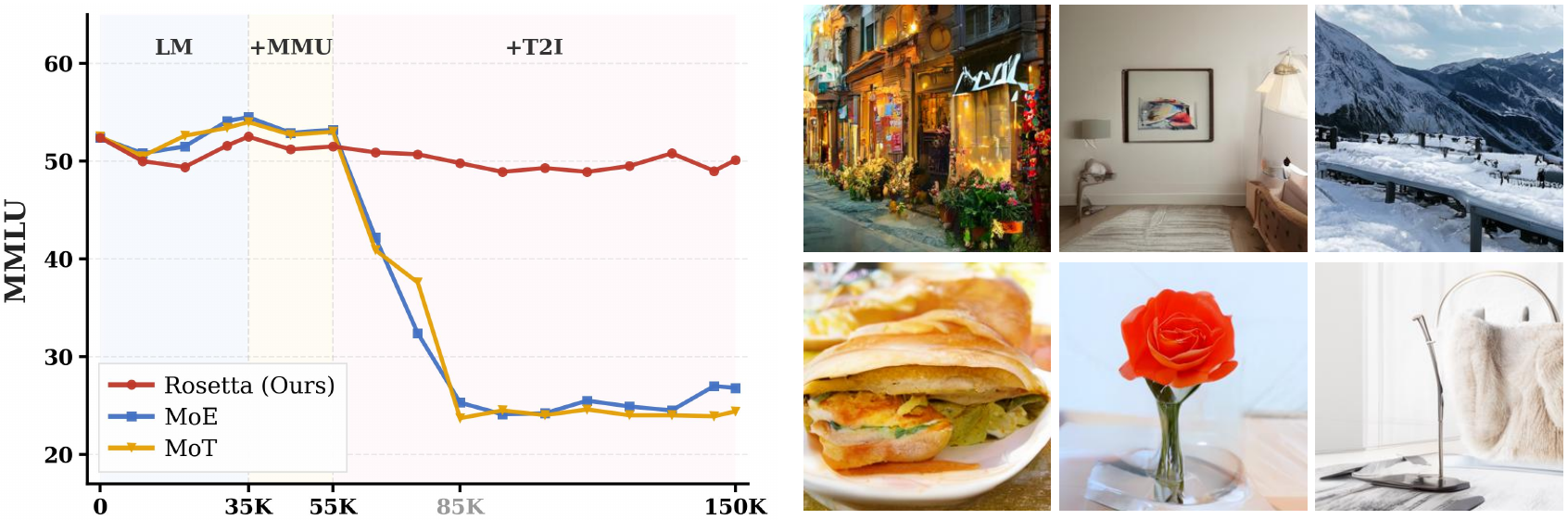}
    % \vspace{-5pt}
    \vspace{-1em} 
    \caption{\textbf{Escaping the Forgetting-Synergy Dilemma.} \textbf{(Left)} Performance dynamics on MMLU benchmark across composable pretraining stages. While standard MoE and structurally isolated MoT suffer from catastrophic routing collapse and degradation upon the integration of continuous generative objectives (+T2I), our Rosetta architecture acts as a robust semantic anchor, maintaining a highly stable foundation. 
    \textbf{(Right)} Qualitative results of Rosetta, demonstrating that the preservation of foundational knowledge seamlessly unlocks superior visual generation capabilities.}
    % high-fidelity 
    \label{fig:teaser}
    \vspace{-1em} 
\end{figure}

The evolution toward general-purpose AI necessitates foundation models capable of natively integrating diverse modalities within a singular architecture~\cite{achiam2023gpt, team2023gemini}, spanning from discrete autoregressive language comprehension to continuous diffusion (or flow matching) visual synthesis~\cite{zhou2024transfusion, xie2024show}. However, integrating these disparate training objectives intrinsically triggers severe gradient conflicts. Specifically, the high-variance gradients from generative tasks tend to overwrite the established representations of language modeling, creating a critical optimization bottleneck.

Scaling unified models effectively points toward Sparse Mixture-of-Experts (MoE)~\cite{jiang2024mixtral, dai2024deepseekmoe}. Yet, standard MoE architectures typically deploy modality-agnostic routing mechanisms. When exposed to heterogeneous multimodal signals, this unconstrained routing leads to a catastrophic \textit{routing collapse}: aggressive generative gradients monopolize and irreversibly overwrite the pre-established experts, severely degrading the model's foundational language (as shown in Fig.~\ref{fig:teaser} Left) and visual understanding capabilities. To mitigate this representation overwriting, structurally partitioned paradigms, such as Mixture-of-Transformers (MoT~\cite{liang2025mixtureoftransformers}) and Bagel~\cite{deng2025emerging}, enforce physical isolation across both the Attention and Feed-Forward Network (FFN) layers within the Transformer~\cite{vaswani2017attention} backbone. While effective at preserving prior knowledge, such rigid structural segregation can inadvertently restrict dense cross-modal interactions, making it challenging to fully leverage cross-modal synergy.

This exposes a critical \textit{Forgetting-Synergy Dilemma}: how can a foundation model effectively expand its generative capabilities without compromising its foundational knowledge, while simultaneously fostering mutual enhancement across modalities? 

In this work, we present Rosetta, a composable native multimodal pretraining framework designed to resolve this dilemma. Much like the historical Rosetta Stone that bridged disparate linguistic scripts, our framework serves as a universal semantic translator, seamlessly harmonizing discrete text, continuous visual perception, and generative latent spaces without mutual interference. Operating on a Lego-like modular paradigm, Rosetta retains a Unified Attention mechanism to preserve high-bandwidth cross-modal contextualization. Concurrently, it confines functional expansion to a composable FFN routing space. By structurally decoupling the FFN into plug-and-play task-specific experts (\eg, dedicated to Text, ViT, or VAE tokens) and a Global Shared Expert, Rosetta isolates task-specific processing while maintaining a universal semantic bridge for cross-modal alignment.

However, the Global Shared Expert inherently remains vulnerable to gradient conflicts. Traditional gradient surgery techniques~\cite{yu2020gradient} require $N$ separate backward passes and gradient buffers for pairwise orthogonalization, incurring an unacceptable $\mathcal{O}(N)$ memory overhead under large-scale distributed training frameworks like FSDP~\cite{zhao2023pytorch}. To overcome this limitation, we propose Momentum-Anchored Orthogonal Projection (MAOP), which innovatively repurposes the optimizer's running momentum state as an implicit semantic anchor. Destructive gradient components from incoming generative tasks are orthogonally projected against this anchor, surgically neutralizing cross-modal interference with strictly zero additional memory overhead.

Our main contributions are summarized as follows:
\begin{itemize}
    \item We propose \textbf{Rosetta}, a composable native multimodal pretraining framework. By structurally decoupling global shared experts from plug-and-play task-specific experts, it seamlessly unifies discrete understanding and continuous generation within a single architecture.
    \item We introduce \textbf{Momentum-Anchored Orthogonal Projection (MAOP)}. To the best of our knowledge, we are the first to leverage optimizer momentum as an implicit semantic anchor to dynamically project gradients, eradicating representation overwriting with  strictly zero additional memory overhead.
    \item Extensive experiments demonstrate that Rosetta eliminates catastrophic forgetting during functional expansion (as in Fig.~\ref{fig:teaser}). Furthermore, it accelerates convergence on new generative tasks and unlocks true cross-modal synergy against prevalent MoE and MoT baselines.
\end{itemize}

\section{Related Work}
\label{sec:related_work}

\vspace{1mm}\noindent\textbf{Unified Multimodal Foundation Models.} Building on the scaling success of foundation models~\cite{achiam2023gpt, team2023gemini, grattafiori2024llama}, recent efforts have focused on integrating diverse modalities into a unified architecture~\cite{alayrac2022flamingo, huang2023language}. Methods such as Chameleon~\cite{team2024chameleon} and Janus~\cite{wu2025janus} typically quantize images into discrete tokens to leverage autoregressive prediction~\cite{van2017neural, razavi2019generating, lee2022autoregressive, lu2024unified, esser2021taming, yu2022scaling}. More recently, paradigms like Transfusion~\cite{zhou2024transfusion} and Show-o~\cite{xie2024show} have demonstrated the superiority of jointly modeling discrete text and continuous visual diffusion~\cite{ho2020denoising, rombach2022high, peebles2023scalable} within a single dense Transformer. However, forcing disparate generative and understanding objectives into a monolithic parameter space inevitably induces severe modality interference, bottlenecking scalability and cross-modal alignment~\cite{niu2025wise}.

\vspace{1mm}\noindent\textbf{Mixture-of-Experts in Multimodal Learning.} Sparse Mixture-of-Experts (MoE) efficiently decouples computation from capacity~\cite{jiang2024mixtral, dai2024deepseekmoe, fedus2022switch, lepikhin2020gshard, riquelme2021scaling}, showing great promise in VLMs~\cite{lin2026moe, li2025uni}. Yet, standard modality-agnostic MoE suffers from catastrophic routing collapse when integrating continuous generation, irreversibly overwriting language experts. To prevent this, recent architectures enforce strict physical separation, such as splitting FFNs~\cite{mckinzie2024mm1, shen2023scaling} or entire Transformer blocks (MoT~\cite{liang2025mixtureoftransformers}, Bagel~\cite{deng2025emerging}). While effectively preventing forgetting, this strict segregation completely severs dense cross-modal synergy. Although shared-expert routing~\cite{liu2026symbiotic} attempts to bridge modalities, it relies on heuristic constraints and lacks the mathematical rigor required for open-ended expansion.

\vspace{1mm}\noindent\textbf{Mitigating Forgetting and Gradient Conflicts.} Expanding foundation models to new modalities frequently triggers catastrophic forgetting due to severe gradient conflicts between disparate objectives. Traditional Continual Learning techniques attempt to preserve prior knowledge via weight regularization~\cite{kirkpatrick2017overcoming, li2017learning, aljundi2018memory} or experience replay~\cite{rebuffi2017icarl, rolnick2019experience, lopez2017gradient, wang2022learning}, but they struggle to scale to billion-parameter distributed pretraining. Alternatively, gradient surgery methods~\cite{yu2020gradient, liu2021conflict, chen2018gradnorm, navon2022multi} tackle this by orthogonally projecting conflicting task gradients. However, this requires materializing separate computational graphs for each task, incurring an unacceptable memory overhead that paralyzes large-scale frameworks~\cite{zhao2023pytorch, rajbhandari2020zero, shoeybi2019megatron, ren2021zero}. Breaking this limitation, our MAOP innovatively repurposes the optimizer's intrinsic momentum as an implicit semantic anchor representing foundational knowledge. By projecting incoming gradients against this anchor, MAOP neutralizes destructive interference with strictly zero additional memory overhead, enabling efficient massive-scale multimodal pretraining.

\section{Method}
\label{sec:method}

\begin{figure}[t!] 
  \centering
  \includegraphics[width=0.99\textwidth]{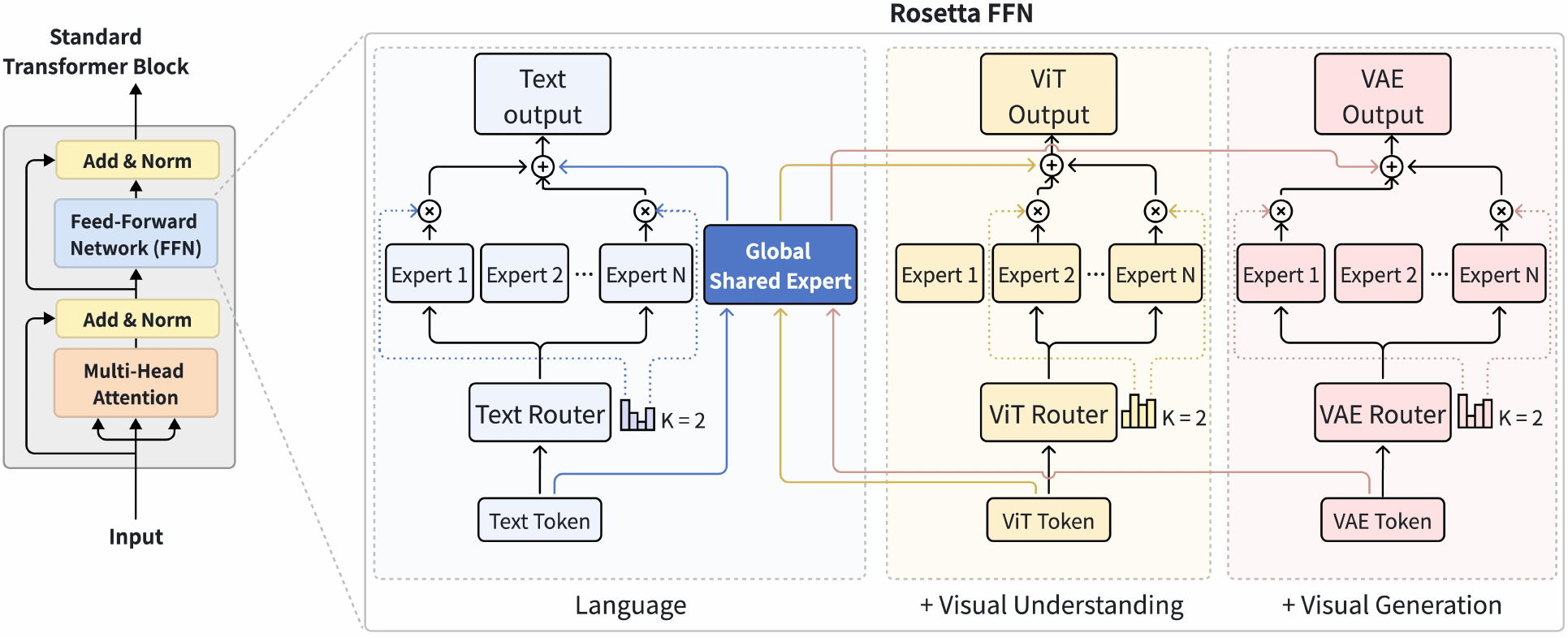}
    \vspace{-5pt}
    \caption{\textbf{Architectural Overview of Rosetta.} Our framework ensures non-destructive modality expansion via three mechanisms: (1) Unified Attention (left): Maintains globally shared QKV projections across all modalities to preserve dense cross-modal interactions. (2) Composable FFN (right): Selectively routes tokens to plug-and-play task-specific experts, bridged by a Global Shared Expert. (3) Conflict-Free Optimization: Momentum-Anchored Orthogonal Projection (MAOP) surgically neutralizes destructive gradients with zero memory overhead, converting modality interference into cross-modal synergy.}
  \label{fig:pipeline}
  \vspace{-10pt}
\end{figure}

In this section, we present Rosetta, a composable native multimodal framework for non-destructive modality expansion. It eliminates catastrophic forgetting and unlocks cross-modal synergy via two core components: (1) \textbf{Rosetta Architecture} (Sec.~\ref{sec:architecture} and Fig.~\ref{fig:pipeline}), which uses Unified Attention and plug-and-play FFN experts linked by a Global Shared Expert; and (2) \textbf{Conflict-Free Optimization} (Sec.~\ref{sec:MAOP}), introducing Momentum-Anchored Orthogonal Projection (MAOP) to neutralize destructive gradients with zero memory overhead. These innovations are operationalized through our \textbf{Composable Pretraining Recipe} (Sec.~\ref{sec:exp_setup} and App.~\ref{app:recipe_stages}), seamlessly harmonizing visual understanding and generation within a native sparse foundation.
% seamlessly integrating visual understanding and generation into a sparse language foundation.

\subsection{Preliminaries}

\vspace{1mm}\noindent\textbf{Standard Mixture-of-Experts (MoE).} 
For an input token $\mathbf{x} \in \mathbb{R}^d$, a standard sparse MoE layer computes the output via a Top-$K$ gating network $\mathcal{G}$:
\begin{equation}
    \mathbf{h}' = \sum_{i=1}^N \mathcal{G}_i(\mathbf{x}) \mathcal{E}_i(\mathbf{x}),
\end{equation}
where $\mathcal{E}_i(\cdot)$ represents the $i$-th expert out of $N$ total experts. Inherent Flaw: Standard routers $\mathcal{G}(\cdot)$ are entirely modality-agnostic. Jointly optimizing heterogeneous signals (\textit{e.g.}, discrete text and continuous vision) under such unconstrained routing triggers severe capacity collapse, irreversibly overwriting pre-established capabilities.

\vspace{1mm}\noindent\textbf{Gradient Conflicts in Multimodal Learning.} 
Expanding a pretrained foundation model to new modalities essentially optimizes a joint objective: $\mathcal{L}_{total} = \mathcal{L}_{base} + \mathcal{L}_{new}$. Let $\mathbf{g}_{base} = \nabla_{\theta} \mathcal{L}_{base}$ and $\mathbf{g}_{new} = \nabla_{\theta} \mathcal{L}_{new}$ represent their respective gradients for shared parameters $\theta$. These objectives frequently diverge, resulting in conflicting gradients where the cosine similarity is negative ($\mathbf{g}_{new}^\top \mathbf{g}_{base} < 0$). Standard optimizers rashly aggregate them ($\mathbf{g}_{total} = \mathbf{g}_{base} + \mathbf{g}_{new}$), pulling shared parameters in opposing directions. This destructive interference is the fundamental optimization root of catastrophic forgetting, necessitating a robust gradient projection mechanism.

\subsection{Rosetta Architecture}
\label{sec:architecture}

As illustrated in Fig.~\ref{fig:pipeline}, Rosetta Transformer block adopts a hybrid paradigm to balance dense cross-modal alignment with interference-free expansion. Specifically, we maintain completely shared QKV projections for dense interactions, while utilizing modality-specific composable sparse FFN layer.

\subsubsection{Unified Attention}
Unlike approaches that enforce early structural isolation via modality-specific QKV projections (\textit{e.g.}, Bagel~\cite{deng2025emerging}, MoT~\cite{liang2025mixtureoftransformers}), Rosetta retains a strictly unified Multi-Head Attention (MHA). Applying modality-specific QKV forces tokens into disjoint representational subspaces prematurely, disrupting global contextualization. By unifying the entire attention operation, Rosetta ensures all tokens are projected into a cohesive semantic space, fostering dense cross-modal interactions prior to modality-aware FFN routing.

\subsubsection{Composable FFN}
While the MHA layer captures dense contextual dependencies, previous studies~\cite{geva2021transformer,meng2022locating} suggest that the Feed-Forward Network (FFN) acts as the primary knowledge repository of the Transformer. Consequently, it is within the FFN that heterogeneous multimodal objectives (\eg, discrete token prediction and continuous diffusion) intrinsically conflict, causing representation overwriting. To fundamentally resolve this bottleneck, Rosetta strictly confines modality-aware expansion to the FFN layer through a highly composable dual-mechanism design.

\vspace{1mm}\noindent\textbf{Modality-Aware Routing with Plug-and-Play Experts.} 
To solve the catastrophic routing collapse observed in standard MoE, Rosetta completely decouples the routing topology. Let $\mathbf{x}^{(t)}$ denote an input token belonging to a specific functional type $t \in \{\text{Text}, \text{ViT}, \text{VAE}\}$. Instead of a single, modality-agnostic router, Rosetta employs modality-specific routers $\mathcal{G}_t(\cdot)$ that restrict token assignment exclusively to a dedicated pool of plug-and-play experts $\{\mathcal{E}_{t, i}\}_{i=1}^{N}$. This explicit decoupling guarantees that high-frequency, noisy generative gradients cannot monopolize the capacity of language or visual understanding experts. Furthermore, this design is natively \textit{composable}: integrating a novel functionality only requires mounting a new router and its corresponding expert group, ensuring extensibility without compromising previously acquired foundational capabilities.

\vspace{1mm}\noindent\textbf{Global Shared Expert as a Cross-Modal Semantic Bridge.} 
Complementing the modality-aware routing against destructive interference, the Global Shared Expert ($\mathcal{E}_{shared}$) serves as the central engine for cross-modal interactions. Rosetta mandates that every token, regardless of its functional type $t$, is deterministically processed by this shared expert. The final FFN output for token $\mathbf{x}^{(t)}$ is formulated as:
\begin{equation}
    \mathbf{h}' = \mathcal{E}_{shared}(\mathbf{x}^{(t)}) + \sum_{i \in \text{TopK}(\mathcal{G}_t(\mathbf{x}^{(t)}))} g_{t, i} \, \mathcal{E}_{t, i}(\mathbf{x}^{(t)}),
\end{equation}
where $g_{t, i}$ represents the routing probability. By absorbing gradients from all diverse tasks, the shared expert learns a universal, modality-agnostic semantic representation. It functions as a global semantic bridge, allowing fine-grained visual generative signals to implicitly enrich language representations, thereby unlocking mutually reinforcing cross-modal synergy.

\subsection{Conflict-Free Optimization via MAOP}
\label{sec:MAOP}

\begin{wrapfigure}{R}{0.4\textwidth}
  \vspace{-2em} % 向上缩进，消除多余白边（顶会省版面神技）
  \centering
  \includegraphics[width=0.33\textwidth]{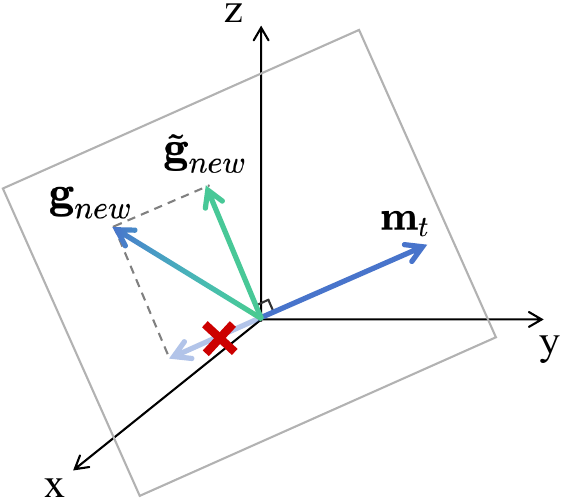}
  \vspace{-1em} % 向下缩进，让caption紧凑
  \caption{\textbf{Illustration of MAOP.}}
  \label{fig:MAOP}
  \vspace{-1.5em} % 向下缩进，让下方文字紧凑
\end{wrapfigure}

While the Rosetta architecture physically isolates modality-specific capabilities, the Global Shared Expert inevitably absorb gradients from all active tasks. When introducing continuous visual generation tasks alongside discrete understanding, the severe heterogeneity of the loss landscapes frequently results in gradient conflicts (\textit{i.e.}, $\mathbf{g}_{new}^\top \mathbf{g}_{base} < 0$). Traditional gradient surgery methods (\textit{e.g.}, PCGrad~\cite{yu2020gradient}) require instantiating and storing separate computational graphs for each task, introducing a prohibitive $\mathcal{O}(N)$ memory overhead that is fundamentally incompatible with large-scale distributed training frameworks like FSDP~\cite{zhao2023pytorch}.

To achieve non-destructive functional expansion, we introduce Momentum-Anchored Orthogonal Projection (MAOP). MAOP innovatively repurposes the optimizer's running momentum $\mathbf{m}_t$ as an \textit{implicit semantic anchor} tracking foundational knowledge. If a conflict is detected ($\mathbf{g}_{new}^\top \mathbf{m}_t < 0$), MAOP surgically neutralizes the interfering component by projecting $\mathbf{g}_{new}$ onto the normal plane of $\mathbf{m}_t$:
\begin{equation}
    \tilde{\mathbf{g}}_{new} = \mathbf{g}_{new} - \frac{\mathbf{g}_{new}^\top \mathbf{m}_t}{\|\mathbf{m}_t\|^2} \mathbf{m}_t, \quad \text{if} \quad \mathbf{g}_{new}^\top \mathbf{m}_t < 0.
\end{equation}
For numerical stability, this projection is bypassed if $\|\mathbf{m}_t\|^2 < 10^{-12}$. Crucially, since $\mathbf{m}_t$ is inherently maintained by the optimizer, MAOP eradicates representation overwriting with strictly zero additional memory overhead, safeguarding synergistic cross-modal updates (details in App.~\ref{app:algorithms}).

\section{Experiments}
\label{sec:exp}

In this section, we comprehensively evaluate Rosetta to answer four core research questions: (1) Can Rosetta natively integrate generative modalities without catastrophic forgetting? (2) Does it unlock true cross-modal synergy compared to physically isolated paradigms? (3) What are the underlying reasons that trigger catastrophic forgetting in standard architectures during expansion? and (4) How do our proposed architectural components (the Global Shared Expert and MAOP) guarantee non-destructive modality expansion? 

\subsection{Experimental Setup}
\label{sec:exp_setup}

\vspace{1mm}\noindent\textbf{Baseline Architectures \& Active Parity.} 
To isolate architectural advancements, all models are upcycled from Qwen3-0.6B Base~\cite{yang2025qwen3} and trained from scratch. We guarantee strict \textit{active parameter parity} ($\sim$0.97B) by mathematically constraining every framework to activate exactly 3 experts (2 routed + 1 shared) per token: 
(1) \textbf{Standard MoE} (denoted simply as MoE): A modality-agnostic baseline routing to 2 out of 12 experts plus 1 shared expert. 
(2) \textbf{MoT}: Representing structural isolation, we instantiate this following Bagel~\cite{deng2025emerging}. It employs modality-specific QKV projections and dual streams (understanding routes to 2 of 7 experts; generation to 2 of 6), each with 1 isolated shared expert. 
(3) \textbf{Rosetta (Ours)}: Maintains unified attention and strictly routes tokens to dedicated task-aware expert pools (3 Text, 3 ViT, 6 VAE), all bridged by a single Global Shared Expert. 
Notably, MoT's structural redundancy inflates its total parameter budget to 4.48B. In contrast, Rosetta preserves the exact 3.77B total parameter of standard MoE while unlocking superior cross-modal synergy (details in App.~\ref{app:param_parity}).

\vspace{1mm}\noindent\textbf{Composable Pretraining Recipe.}
% For strict fairness, all models process identical data sequences in pretraining via fixed seeds and without any subsequent scaling or tuning stages.
For strict fairness, all models process identical pretraining sequences via fixed seeds, without any subsequent scaling or tuning stages.
% downstream instruction tuning. 
% For strict fairness, all methods process identical training data sequences via fixed seeds in pretraining setting, without any subsequent scaling or tuning stages.
% For strict fairness and to isolate architectural impacts, all models process identical data sequences in training via fixed seeds and are evaluated in their raw pretraining foundation states without downstream instruction tuning. 
The vision encoder Qwen3-VL ViT~\cite{bai2025qwen3} and FLUX.2 VAE~\cite{flux2_2025} remain permanently frozen in all settings. Pretraining spans three composable stages: (1) \textit{Language Foundation (LM)}: Models are optimized on $\sim$300B text tokens for 35K steps. (2) \textit{Visual Understanding (+MMU)}: A 3K-step LLaVA-style projector warmup~\cite{liu2023visual} precedes 20K joint training steps ($\sim$4M MMU samples, MMU:LM = 0.8:0.2). (3) \textit{Visual Generation (+T2I)}: Joint optimization of all modalities using a data sampling ratio of T2I:MMU:LM = 0.6:0.25:0.15. Comprehensive recipes are in App.~\ref{app:recipe_stages} and hyperparameters in App.~\ref{app:hyperparams}.

\begin{table*}[t]
\centering
\vspace{-1em}
\caption{\textbf{Comprehensive Performance Evaluations.} 
All methods are evaluated in their raw foundation state after the full multimodal pretraining phase (LM+MMU+T2I) under identical training constraints, strictly without any downstream instruction tuning. 
Rosetta fundamentally overcomes the catastrophic forgetting existing in MoE and MoT, achieving the best across all three capability domains (Language, Visual Understanding, and Visual Generation). The gray row Rosetta (Pre-T2I) represents the understanding performance of Rosetta prior integrating T2I. (\textbf{Bold}: Best)}
\label{tab:main_results}

% 魔法指令 0：强行撑开 Caption 和表格的距离！让辛金的排版呼吸起来！
\vspace{5pt} % 如果觉得还不够，可以改成 10pt 或者 2mm、3mm

% 魔法指令 1：强制缩小列间距，压榨空白，瞬间放大全局字体！
\setlength{\tabcolsep}{1.0pt} 

\resizebox{\textwidth}{!}{%
\begin{tabular}{l | ccc | ccccccc}
\toprule
\multirow{2}{*}{\textbf{Method}} & \textbf{Total} & \textbf{Active} & \textbf{Training} & \multicolumn{7}{c}{\textbf{Visual Generation}} \\
\cmidrule{5-11}
 & \textbf{Params} & \textbf{Params} & \textbf{Iterations} & \textbf{T2I-Comp}$\uparrow$ & \textbf{Color}$\uparrow$ & \textbf{Shape}$\uparrow$ & \textbf{Texture}$\uparrow$ & \textbf{FID}$\downarrow$ & \textbf{CLIPScore}$\uparrow$ & \textbf{HPSv2}$\uparrow$ \\
\midrule
MoE & 3.77B & 0.97B & 400K & \text{40.2} & \text{47.7} & \text{27.5} & \text{45.5} & \text{17.80} & \text{0.287} & \text{0.204} \\
MoT (Bagel) & 4.48B & 0.97B & 400K & \text{43.5} & \text{50.7} & \text{29.8} & \text{50.0} & \text{15.58} & \text{0.288} & \text{0.211} \\
\rowcolor{gray!10} % 尊贵的浅灰色高亮
Rosetta & 3.77B & 0.97B & 400K & \textbf{45.5} & \textbf{52.9} & \textbf{31.7} & \textbf{51.9} & \textbf{14.05} & \textbf{0.290} & \textbf{0.219} \\

\midrule\midrule 

\multirow{2}{*}{\textbf{Method}} & \multicolumn{4}{c|}{\textbf{Language}} & \multicolumn{6}{c}{\textbf{Visual Understanding}} \\
\cmidrule{2-11}
 & \textbf{MMLU}$\uparrow$ & \textbf{BBH}$\uparrow$ & \multicolumn{1}{c}{\textbf{ARC-c}$\uparrow$} & \multicolumn{1}{c|}{\textbf{MBPP}$\uparrow$} & \textbf{MMMU}$\uparrow$ & \textbf{MMB-EN}$\uparrow$ & \textbf{MMB-CN}$\uparrow$ & \textbf{POPE}$\uparrow$ & \textbf{AI2D}$\uparrow$ & \textbf{RealWorldQA}$\uparrow$ \\
\midrule
\textcolor{gray}{Rosetta (Pre-T2I)} & \textcolor{gray}{51.6} & \textcolor{gray}{48.8} & \multicolumn{1}{c}{\textcolor{gray}{67.3}} & \multicolumn{1}{c|}{\textcolor{gray}{36.2}} & \textcolor{gray}{34.1} & \textcolor{gray}{51.3} & \textcolor{gray}{45.2} & \textcolor{gray}{78.1} & \textcolor{gray}{54.2} & \textcolor{gray}{47.4} \\
\midrule
MoE & \text{26.3} & \text{0} & \multicolumn{1}{c}{\text{22.8}} & \multicolumn{1}{c|}{\text{0}} & \text{26.8} & \text{42.1} & \text{31.6} & \text{77.0} & \text{41.6} & \text{39.5} \\
MoT (Bagel) & \text{27.1} & \text{0} & \multicolumn{1}{c}{\text{26.5}} & \multicolumn{1}{c|}{\text{0}} & \text{27.1} & \text{46.2} & \text{40.7} & \text{74.0} & \text{47.6} & \text{45.1} \\
\rowcolor{gray!10} % 尊贵的浅灰色高亮
Rosetta & \textbf{49.2} & \textbf{46.8} & \multicolumn{1}{c}{\textbf{62.9}} & \multicolumn{1}{c|}{\textbf{42.4}} & \textbf{34.6} & \textbf{52.5} & \textbf{45.7} & \textbf{80.1} & \textbf{55.8} & \textbf{48.2} \\

\bottomrule
\end{tabular}%
}
 \vspace{-1.5em} 
\end{table*}

\vspace{1mm}\noindent\textbf{Evaluation Metrics.} 
To comprehensively evaluate models, we conduct experiments across three main domains. For \textit{Language}, we measure general knowledge and reasoning via MMLU~\cite{hendryckstest2021} and BBH~\cite{suzgun2023challenging}, language understanding via ARC-challenge~\cite{clark2018think}, and coding capabilities via MBPP~\cite{austin2021program}. For \textit{Visual Understanding}, we utilize a diverse set of tests: MMMU~\cite{yue2024mmmu} (expert-level reasoning), MMBench (English and Chinese)~\cite{liu2024mmbench} (comprehensive perception), POPE~\cite{li2023evaluating} (hallucination robustness), AI2D~\cite{kembhavi2016diagram} (diagram understanding), and RealWorldQA~\cite{realworldqa2024} (real-world comprehension). For \textit{Visual Generation}, synthesis fidelity and semantic alignment are evaluated using FID~\cite{heusel2017gans}, CLIPScore~\cite{hessel2021clipscore}, and HPSv2~\cite{wu2023human} on COCO-30K~\cite{lin2014microsoft}. Furthermore, we employ T2I-CompBench~\cite{huang2023t2i} to explicitly quantify compositional prompt adherence across specific attributes (\textit{i.e.}, Color, Shape, and Texture).

\begin{figure}[t!] 
  \centering
  \vspace{-1em} 
  \includegraphics[width=1.0\textwidth]{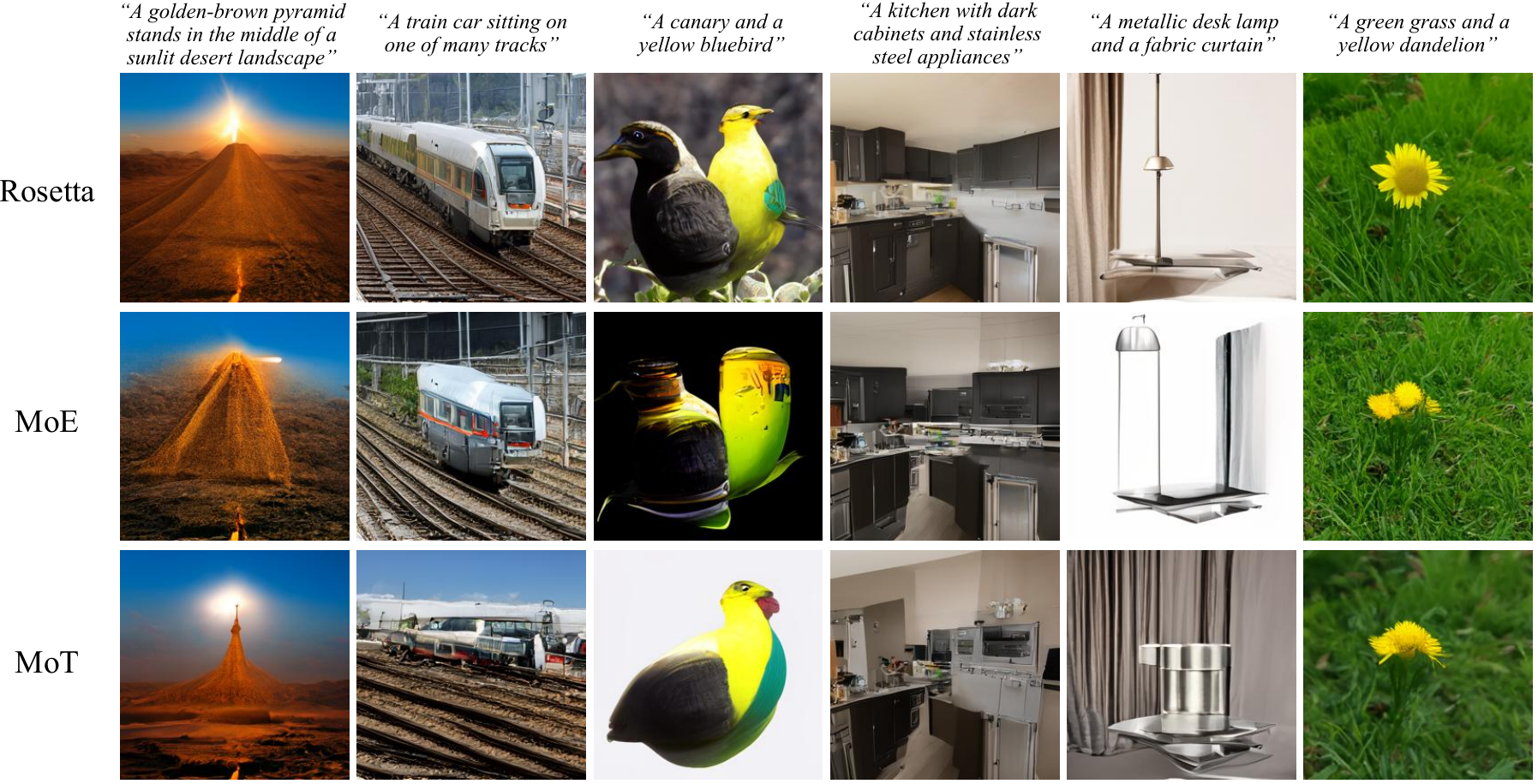}
   \vspace{-5pt}
    % \caption{\textbf{Visual Quality}}
    \vspace{-1em} 
    \caption{\textbf{Qualitative Comparisons.} Standard MoE suffers semantic drift (\textit{e.g.}, bird to bottle) and MoT exhibits structural distortions (\textit{e.g.}, broken lamp). In contrast, Rosetta leverages cross-modal synergy to synthesize high-fidelity images with precise spatial geometry and prompt adherence.} 
  \label{fig:exp_mix}
% \vspace{-10pt}
\vspace{-1em} 
\end{figure}

\subsection{Main Results: Escaping the Forgetting-Synergy Dilemma}
\label{sec:exp_main}

\begin{figure}[t!]
    \centering
    \vspace{-1.5em} 
    % === 第一行：宏观稳定性 (Macro Dynamics) ===
    \includegraphics[width=0.32\textwidth]{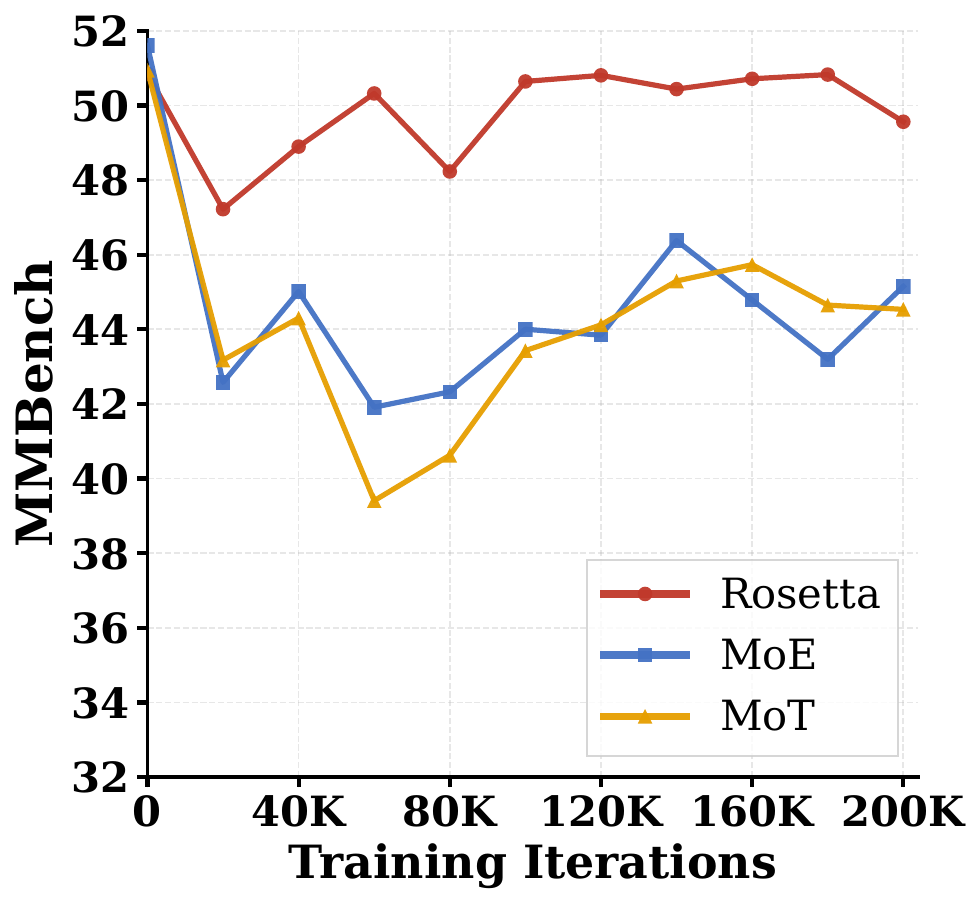}\hfill
    \includegraphics[width=0.31\textwidth]{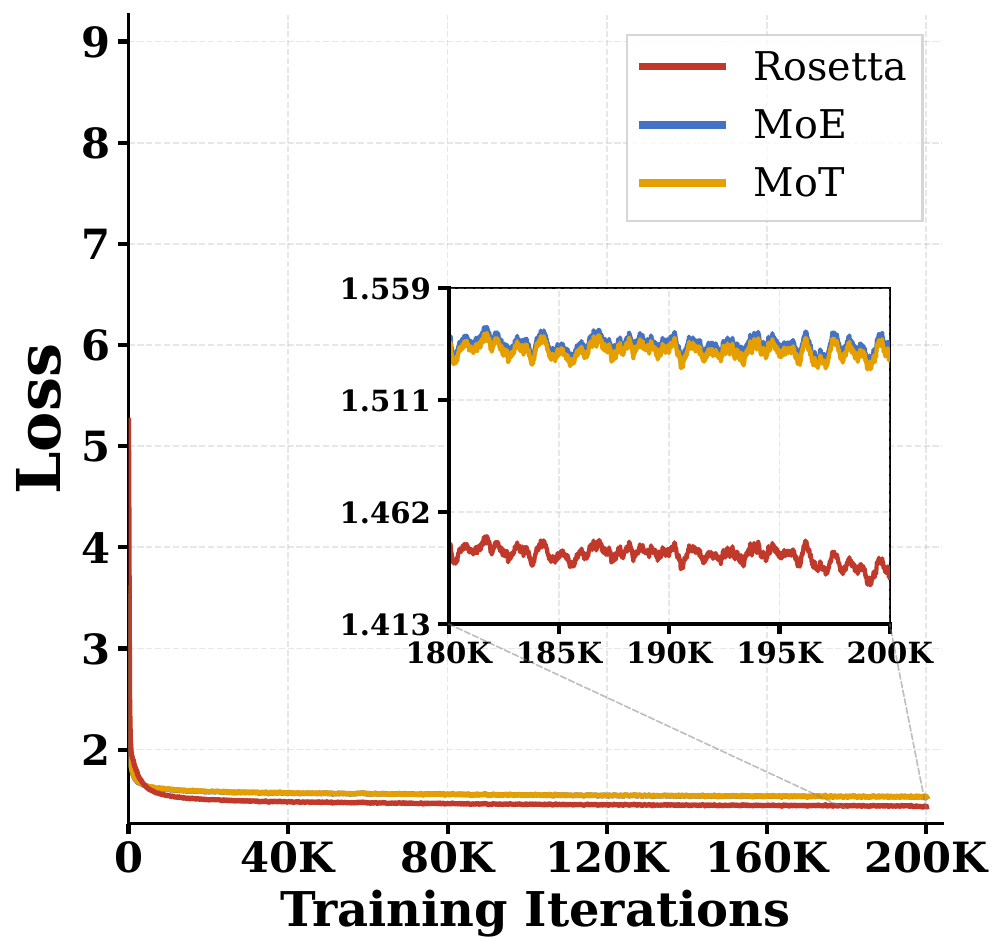}\hfill
    \includegraphics[width=0.33\textwidth]{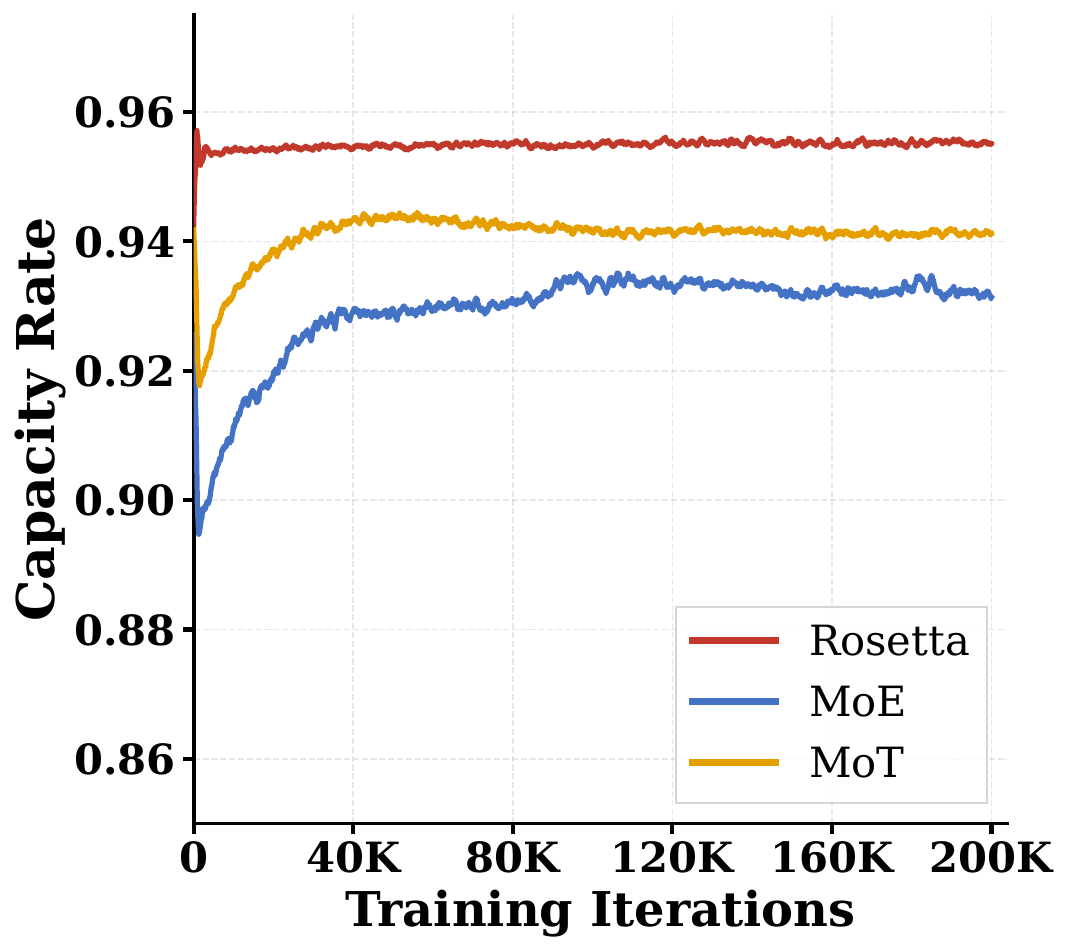}
    \vspace{2mm} % 两行之间的微小间距，保持呼吸感
    % === 第二行：微观 Loss 解剖 (Micro-Loss Decomposition) ===
    \includegraphics[width=0.32\textwidth]{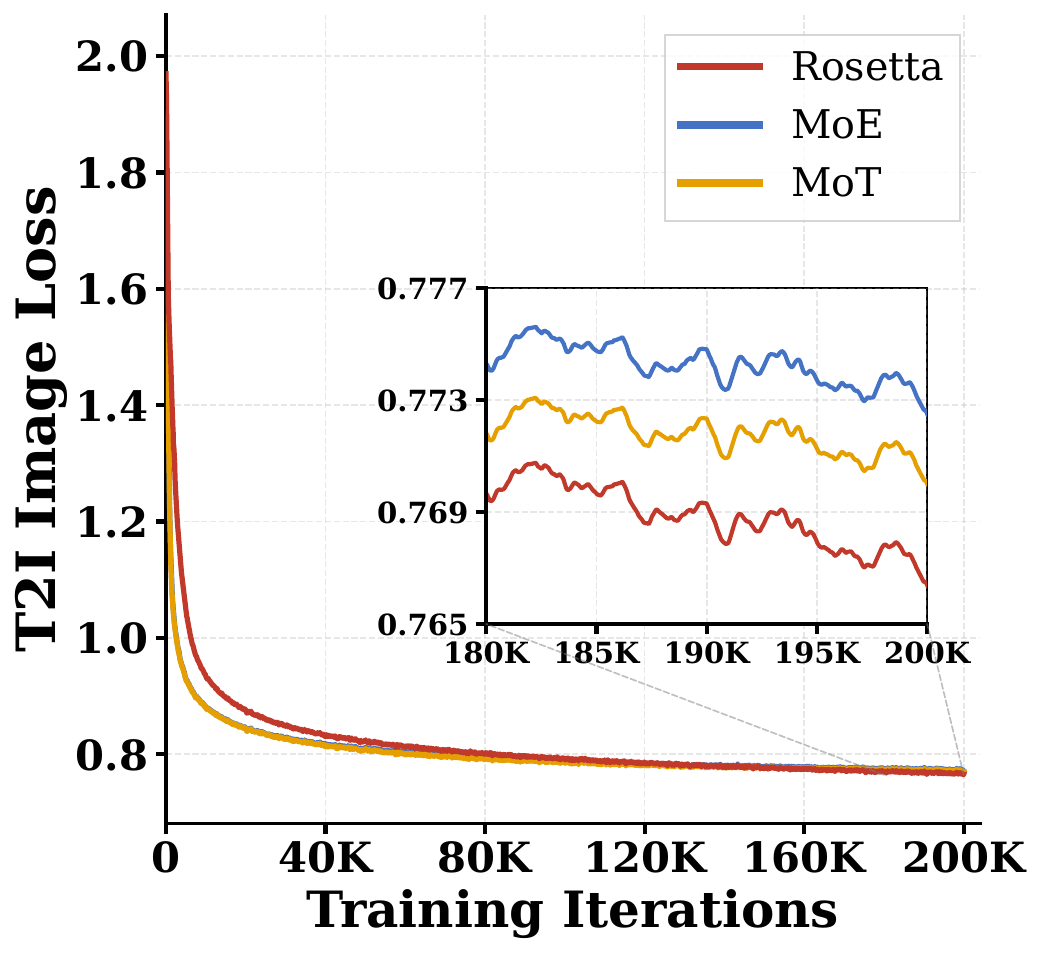}\hfill
    \includegraphics[width=0.32\textwidth]{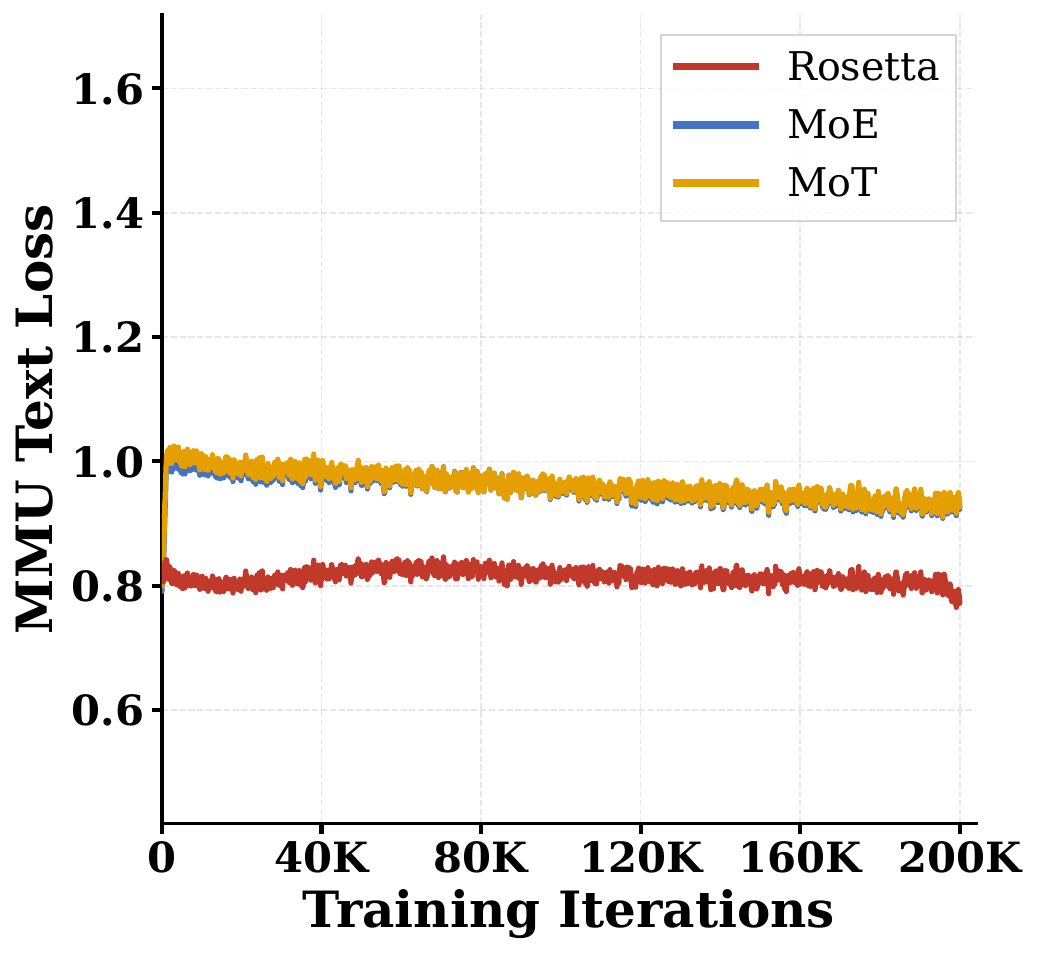}\hfill
    \includegraphics[width=0.33\textwidth]{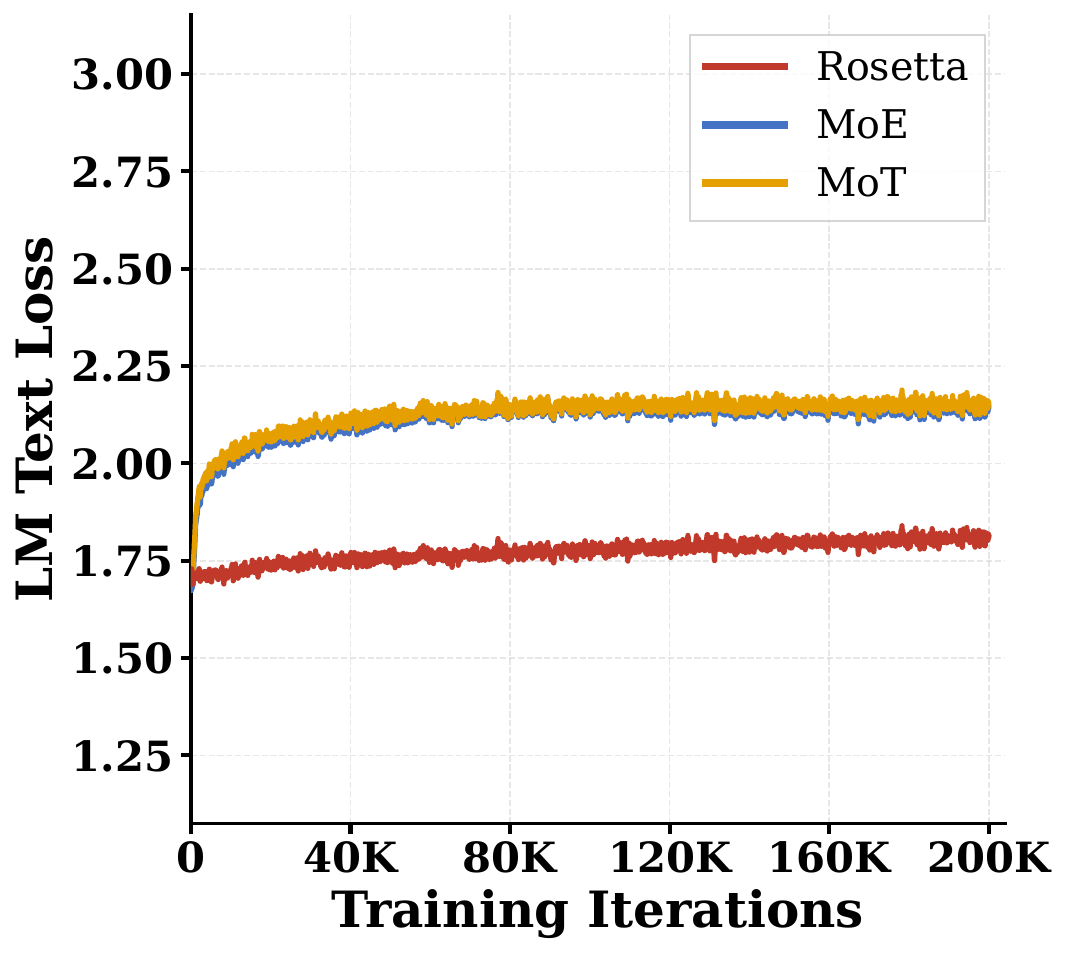}
    % \vspace{-5pt}
    \vspace{-1em} 
    \caption{\textbf{Comprehensive Training Dynamics.} Evaluated over a 200K-step generative expansion. \textbf{(1) Overall Dynamics (Top Row):} Rosetta averts the irreversible MMBench degradation in MoE and MoT baselines, maintaining a synergistic upward trajectory (Left). It also achieves a deeper optimization bound (Center) and near-optimal capacity rate (\textit{i.e.}, ratio of successfully routed, non-dropped tokens; $\sim$0.95, Right). \textbf{(2) Task-Specific Losses (Bottom Row):} Rosetta accelerates T2I convergence (Left) and neutralizes cross-modal gradient interference, guaranteeing strictly lower and stable trajectories for both visual (Center) and language understanding (Right).}
    \label{fig:comprehensive_dynamics}
    % \vspace{-15pt} % 压缩图注下方的白边，极其省空间
    \vspace{-1.5em} 
\end{figure}

\vspace{1mm}\noindent\textbf{Language Ability.} 
As evidenced by the MMLU dynamics in Fig.~\ref{fig:teaser}, integrating image generation triggers a severe routing collapse in baseline architectures. The underlying cause is revealed in Fig.~\ref{fig:comprehensive_dynamics} (bottom right): the aggressive injection of T2I gradients causes the LM Text Loss of MoE and MoT to catastrophically diverge. This optimization instability translates to massive performance degradation across all language metrics (as in Tab.~\ref{tab:main_results}). In stark contrast, Rosetta effectively absorbs these generative shocks, maintaining a strictly suppressed and stable loss trajectory. 
Benefiting from this architectural immunity, Rosetta preserves foundational language capabilities with minimal degradation compared to its pre-generative checkpoint Pre-T2I (trained exclusively on language and visual understanding). This stark divergence is most vividly illustrated in highly sensitive reasoning (BBH) and coding (MBPP) benchmarks. While generative interference completely destroys these capabilities in MoE and MoT—collapsing both their BBH and MBPP scores to 0, Rosetta robustly safeguards complex reasoning (BBH retaining 46.8 vs. 48.8) and even enhances coding proficiency (MBPP increasing from 36.2 to 42.4), demonstrating exceptional cross-modal resilience.

\vspace{1mm}\noindent\textbf{Visual Understanding.} 
Building upon its stable language capabilities, Rosetta further improves visual understanding through cross-modal synergy. As detailed in Tab.~\ref{tab:main_results}, the integration of generative tasks causes a universal performance drop across all visual understanding metrics for both MoE and MoT. Although attempts like freezing the understanding branch in MoT (\textit{e.g.}, Bagel) avoids interference, but strictly limits visual understanding performance to pre-trained levels and prevents further enhancement. In contrast, Rosetta achieves consistent improvements over its Pre-T2I baseline across all indicators. This superiority directly stems from the underlying optimization dynamics (Fig.~\ref{fig:comprehensive_dynamics}, bottom center): while the MMU Text Loss of baselines severely diverges under generative interference, Rosetta maintains a highly stable trajectory. Consequently, as tracked by the MMBench dynamics (Fig.~\ref{fig:comprehensive_dynamics}, top left), although all methods experience an initial performance drop, MoE and MoT suffer irreversible degradation. Rosetta, however, rapidly recovers and maintains a steady upward trend. This confirms that Rosetta successfully unlocks true cross-modal synergistic evolution.

\vspace{1mm}\noindent\textbf{Visual Generation.} 
Crucially, Rosetta's preservation of foundational knowledge does not compromise generative plasticity. As detailed in Tab.~\ref{tab:main_results}, Rosetta consistently achieves the best performance across all visual generation metrics, delivering superior synthesis fidelity on COCO-30K and dominating compositional benchmarks like T2I-CompBench. This empirical success is directly supported by its optimization trajectory (Fig.~\ref{fig:comprehensive_dynamics}, bottom left). Rosetta achieves the lowest T2I Image Loss in training, this optimization efficacy stems from robust cross-modal synergy, effectively resolving parameter competition and enabling the model to achieve superior generative performance. Qualitative comparisons (Fig.~\ref{fig:exp_mix}) further validate that unlike MoE and MoT which suffer semantic drift or structural distortions, Rosetta synthesizes high-fidelity images with precise prompt adherence, successfully dismantling the traditional stability-plasticity dilemma.

\vspace{1mm}\noindent\textbf{Unlocking Cross-Modal Synergy.} In summary, Rosetta effectively overcomes the forgetting-synergy dilemma. Upon integrating continuous generative tasks, MoE and MoT suffer catastrophic degradation across all language and visual understanding metrics. Conversely, Rosetta robustly preserves foundational language priors with even improving coding proficiency, and universally enhances visual understanding performance. Coupled with significantly faster convergence in visual generation, Rosetta demonstrates that introducing new modalities can serve as a constructive regularizer rather than a disruptive force. This mutually beneficial synergy establishes a highly scalable and unified blueprint for extensible multimodal foundation models.

% \vspace{-1.5cm} % 【微操指令】：如果吸得不够，就改成 -2.0cm！如果撞上了，就改成 -1.0cm！

\subsection{In-depth Analysis and Ablation Studies}
\label{sec:exp_ablation}

\begin{figure}[t!]
    \centering
    \vspace{-1.0em} % 【军师微操1】：如果页眉有空隙，强行把图往上顶
    \includegraphics[width=\textwidth]{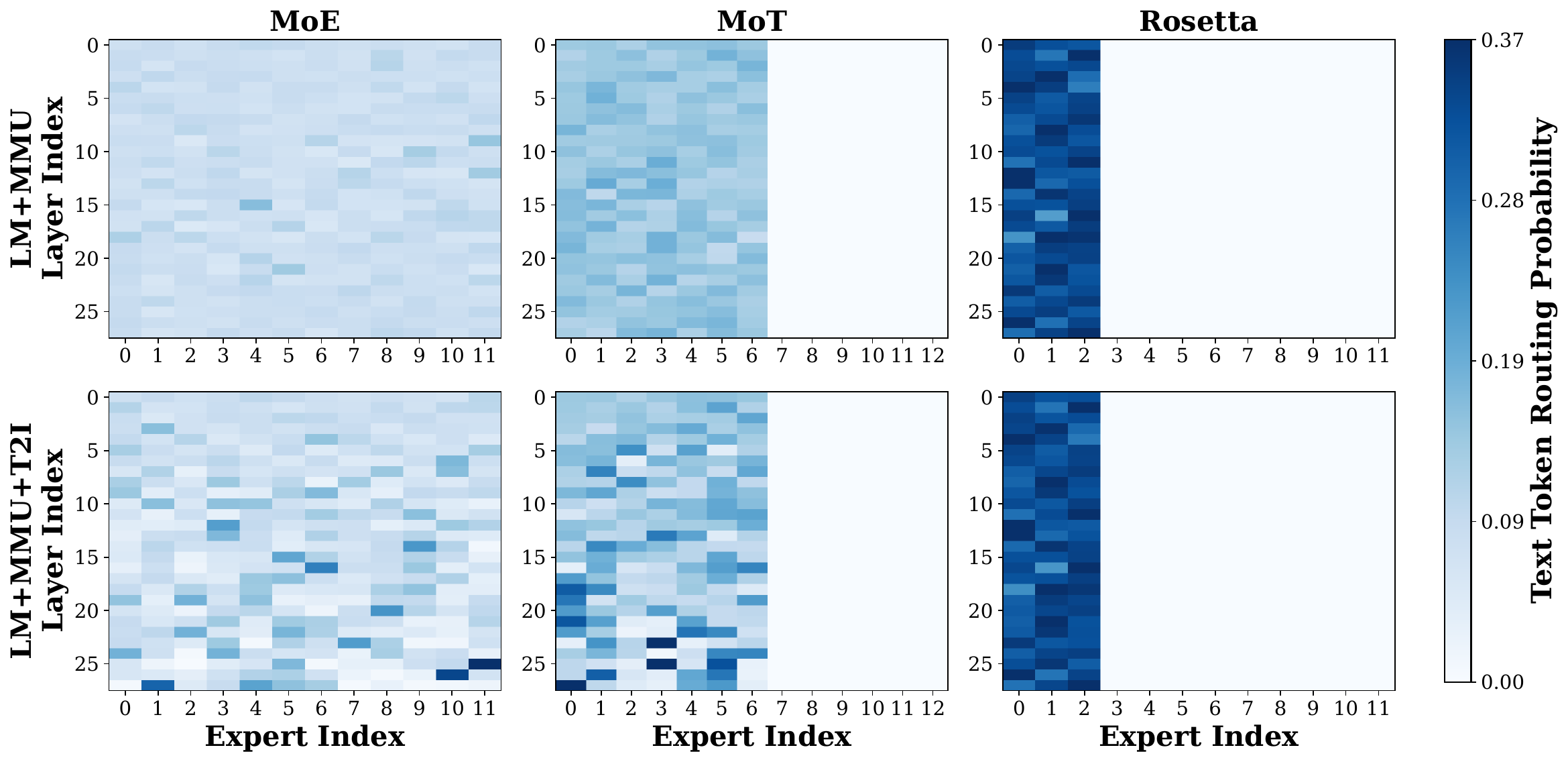}
    \vspace{-1.5em} % 【军师微操2】：把 Caption 强行往图片方向吸，消除图与字的裂缝
    \caption{\textbf{Routing Distribution Heatmaps During Generative Expansion.} We visualize the routing probabilities of Text tokens across experts during MMLU inference. \textbf{Top Row:} Checkpoints under the LM+MMU configuration (iteration 55K in Fig.~\ref{fig:teaser}). \textbf{Bottom Row:} Checkpoints upon integrating 30K steps of T2I training (iteration 85K in Fig.~\ref{fig:teaser}). Both MoE and MoT exhibit significant distribution shifts, indicating severe cross-modal interference. In contrast, Rosetta maintains nearly identical routing distribution, successfully preserving pre-established language capabilities.}
    \label{fig:routing_heatmaps}
    \vspace{-1.5em} % 【军师绝杀3】：强行把下方的正文往上吸！彻底消灭死亡白边！
\end{figure}

\vspace{1mm}\noindent\textbf{Deep Analysis: Unmasking the Collapse.} 
To investigate the severe MMLU performance drop observed in Fig.~\ref{fig:teaser}, we visualize the routing distribution of Text tokens during inference. Fig.~\ref{fig:routing_heatmaps} compares the expert activation probabilities at two critical checkpoints from Fig.~\ref{fig:teaser}: before generative training (top row, iteration 55K) and after 30K steps of adding T2I training (bottom row, iteration 85K). 
For MoE (Left), the text routing distribution shifts significantly, indicating that continuous generative gradients severely interfere with the pre-trained modality-agnostic experts. Notably, MoT (Middle) also exhibits clear routing changes despite having physically separated experts. This reveals that structural isolation alone is insufficient, generative signals can still corrupt shared parameters during joint optimization. In contrast, Rosetta (Right) maintains nearly the same routing distribution before and after the generative expansion. By combining modality-aware routing with the MAOP mechanism, Rosetta effectively eliminates cross-modal interference and preserves pre-established capabilities.

% 环绕表
\begin{wraptable}{R}{0.50\textwidth} 
  \vspace{-1.5em} % 强行上移，抹平白边
% 魔法指令 1：强制缩小列间距，压榨空白，瞬间放大全局字体！
\setlength{\tabcolsep}{2pt} 
  \centering
  \caption{\textbf{Ablation Results.} Impact of removing core components on multimodal synergy.}
  \label{tab:ablation}
  \vspace{0.3em} % 给 Caption 和表格一点呼吸感
  \resizebox{\linewidth}{!}{
  \begin{tabular}{l c c c c}
  \toprule
  \textbf{Variant} & \textbf{Iters} & \textbf{MMLU}$\uparrow$ & \textbf{MMBench}$\uparrow$ & \textbf{FID}$\downarrow$ \\
  \midrule
  w/o Shared Expert & 100k & 49.84 & 49.19 & 36.13 \\
  w/o MAOP & 100k & 45.48 & 46.07 & 34.29 \\
  \rowcolor{gray!10}
  Rosetta (Full) & 100k & \textbf{50.35} & \textbf{50.64} & \textbf{34.21} \\
  \bottomrule
  \end{tabular}
  }
  \vspace{-1.0em} % 消除底部的死亡留白
\end{wraptable}

\vspace{1mm}\noindent\textbf{Efficacy of the Global Shared Expert.} 
Removing the Global Shared Expert reduces our FFN to a strictly isolated routing paradigm. Notably, unlike MoT, this variant retains unified QKV projections, allowing us to precisely isolate the impact of the FFN's shared representation space. As detailed in Tab.~\ref{tab:ablation}, while this strict isolation within the FFN successfully averts catastrophic forgetting, it incurs a severe penalty in compositional generation tasks (FID) and advanced multimodal reasoning (MMBench). This empirically confirms that rigid isolation inherently impedes cross-modal synergy, establishing our Global Shared Expert as an indispensable semantic bridge for harmonizing diverse modalities.

\vspace{1mm}\noindent\textbf{Necessity of MAOP.} 
To verify our gradient projection mechanism, we train a variant without MAOP. As illustrated in Tab.~\ref{tab:ablation}, this variant experiences a clear degradation in language understanding metrics upon the introduction of visual generation. This indicates that while structural decoupling prevents routing collapse, the Shared Expert still suffers from representation overwriting due to gradient conflicts. MAOP surgically neutralizes this interference.

\vspace{1mm}\noindent\textbf{System-Level Efficiency and Zero Overhead.} Finally, we highlight the architectural elegance of MAOP compared to conventional gradient surgery techniques. Mathematically, PCGrad requires $N$ separate backward passes (one per task, \textit{i.e.}, $N{=}3$ in our LM\,+\,MMU\,+\,T2I setting), reducing training throughput by approximately $3{\times}$, while materializing $N$ distinct gradient tensors introduces the same $\mathcal{O}(N)$ memory overhead described above. Conversely, MAOP repurposes the optimizer's intrinsically pre-allocated momentum state, requiring no additional backward passes or gradient buffers. Empirical hardware profiling confirms that enabling MAOP introduces strictly zero additional peak memory (maintaining exactly 58.8\,GB per GPU) and zero computational overhead (retaining an identical throughput of 8{,}602 tokens/s/GPU). This establishes MAOP as an exceptionally viable, zero-cost optimization solution for large-scale foundation models.

\begin{wrapfigure}{R}{0.33\textwidth}
    \vspace{-1.5em} % 【军师微操】：强制上移，对齐段落第一行文字
    \centering
    \includegraphics[width=\linewidth]{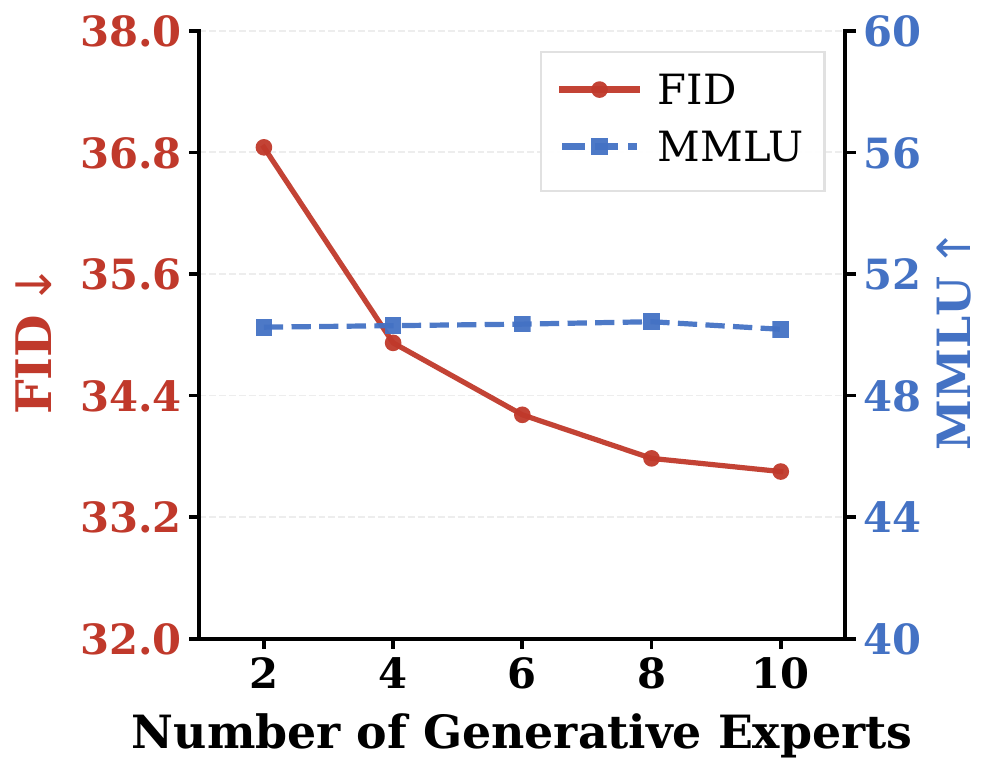}
    \vspace{-2.0em}
    \caption{\textbf{Expert Scalability of Rosetta.} 
    }
    \label{fig:scaling}
    \vspace{-1.0em} % 【军师微操】：切断底部的死亡留白
\end{wrapfigure}
\vspace{1mm}\noindent\textbf{Expert Scalability.} 
We furture analyze the scalability of Rosetta's plug-and-play experts. By varying the number of generation experts ($N_{VAE} \in \{2, 4, 6, 8, 10\}$) while maintaining the active parameter count ($\sim$0.97B) constant, we evaluate the generative fidelity at a 100K-step checkpoint to observe structural scalability. As illustrated in Fig.~\ref{fig:scaling}, expanding the expert pool monotonically improves synthesis fidelity (evidenced by a continuous decrease in FID) without degrading the established language understanding capabilities (MMLU remains rigorously stable). This demonstrates robust expert scaling behavior, confirming Rosetta's potential for highly efficient and non-destructive modality expansion.

\section{Conclusion}
\label{sec:conclusion}

In this work, we present Rosetta to resolve the Forgetting-Synergy Dilemma between discrete understanding and continuous generation. Structurally, it delegates modality-specific processing to plug-and-play experts while employing a Global Shared Expert as a semantic bridge. Mathematically, our Momentum-Anchored Orthogonal Projection (MAOP) neutralizes destructive gradient conflicts with strictly zero additional memory overhead. Ultimately, Rosetta goes beyond a mere architectural improvement; it introduces a scalable philosophy for next-generation foundation models. As the community accelerates toward Artificial General Intelligence, seamlessly integrating new modalities including audio, video, 3D perception, and embodied control becomes paramount.
By mathematically and structurally guaranteeing this non-destructive expansion, Rosetta provides a blueprint for such a future: a truly unified and ever-expanding native multimodal foundation model.

\medskip
{
\small
\bibliographystyle{plainnat}
\bibliography{neurips_2026} 
}

\newpage
\appendix

\begin{center}
{\Large \textbf{Supplementary Materials for \\ Rosetta: Composable Native Multimodal Pretraining}}
\end{center}
\vspace{0.5cm}

This supplementary document provides comprehensive theoretical foundations, system-level implementation specifics, extended empirical configurations, and additional visualizations to guarantee the absolute reproducibility of our framework. The contents are systematically organized as follows:

\begin{itemize}
    \item \textbf{Appendix~\ref{app:algorithms}: Algorithmic Foundations and Mathematical Proofs.} Provides the formal formulation of Momentum-Anchored Orthogonal Projection (MAOP), proves its strict mathematical correctness and zero-overhead scalability under FSDP, and details the magnitude-preserving sparse upcycling strategy.
    
    \item \textbf{Appendix~\ref{app:implementation}: System-Level Implementation Details.} Elaborates on the two-stage synergistic protection mechanism and the Differential Learning Rate (DiffLR) strategy, providing explicit pseudo-code for its native optimization overriding within distributed frameworks.
    
    \item \textbf{Appendix~\ref{app:curriculum}: Comprehensive Pretraining Curriculum and Configurations.} Presents the detailed modality expansion pipeline, comprehensive hyperparameter settings, data sampling ratios, and the exact architectural parameter breakdown that rigorously ensures active computational parity.
    
    \item \textbf{Appendix~\ref{app:qualitative}: Additional Qualitative Results.} More image generation results that corroborate Rosetta's superior cross-modal alignment.
\end{itemize}
\vspace{0.3cm}

% ==========================================
% App. A: 算法与底层数学证明 (秀肌肉专区)
% ==========================================
\section{Algorithmic Foundations and Mathematical Proofs}
\label{app:algorithms}

\subsection{Momentum-Anchored Orthogonal Projection (MAOP)}
\label{app:maop_math}

Let $\mathbf{g} \in \mathbb{R}^D$ denote the current mixed gradient (aggregated from language and visual generative tokens) for the Shared Expert parameters. Let $\mathbf{m} \in \mathbb{R}^D$ denote the optimizer's first-moment estimate (\textit{e.g.}, \texttt{exp\_avg} in AdamW), representing the exponentially moving average of historical gradients. This $\mathbf{m}$ serves as an implicit semantic anchor for the established learning trajectory.

At each backward pass, MAOP evaluates the cosine similarity between the current gradient and the historical momentum. A destructive conflict is defined as their inner product being negative ($\mathbf{g}^\top \mathbf{m} < 0$). To neutralize this interference, MAOP orthogonally projects $\mathbf{g}$ onto the normal plane of $\mathbf{m}$:
\begin{equation}
\mathbf{g}_{\text{orth}} = 
\begin{cases} 
\mathbf{g} - \frac{\mathbf{g}^\top \mathbf{m}}{\|\mathbf{m}\|^2 + \epsilon} \mathbf{m}, & \text{if } \mathbf{g}^\top \mathbf{m} < 0 \\
\mathbf{g}, & \text{otherwise}
\end{cases}
\end{equation}
where $\epsilon$ is a small constant for numerical stability. 

\textbf{Proof of Non-Interference:} To mathematically verify that the projected gradient $\mathbf{g}_{\text{orth}}$ exerts exactly zero interference on the momentum trajectory $\mathbf{m}$, we compute their inner product in the conflict scenario ($\mathbf{g}^\top \mathbf{m} < 0$):
\begin{equation}
    \mathbf{g}_{\text{orth}}^\top \mathbf{m} = \left( \mathbf{g} - \frac{\mathbf{g}^\top \mathbf{m}}{\|\mathbf{m}\|^2} \mathbf{m} \right)^\top \mathbf{m} = \mathbf{g}^\top \mathbf{m} - \frac{\mathbf{g}^\top \mathbf{m}}{\|\mathbf{m}\|^2} (\mathbf{m}^\top \mathbf{m}) = 0.
\end{equation}
This confirms that MAOP completely removes the antagonistic component while preserving synergistic updates ($\mathbf{g}_{\text{orth}}^\top \mathbf{m} \geq 0$), mathematically eradicating representation overwriting.

\subsection{Scalable Distributed Implementation in FSDP}
\label{app:maop_fsdp}

A critical engineering challenge in modern massive-scale pretraining is that model parameters and their corresponding gradients are heavily sharded across multiple GPUs. Computing the global inner product $\mathbf{g}^\top \mathbf{m}$ directly on local shards would mathematically violate the projection geometry.

To maintain rigorous mathematical equivalence with strictly zero memory overhead, MAOP is implemented using a synchronized distributed reduction. Specifically, we first compute the local inner product on each rank's shard: $S_{\text{local}} = \mathbf{g}^{(rank)} \cdot \mathbf{m}^{(rank)}$ and $N_{\text{local}} = \|\mathbf{m}^{(rank)}\|^2$. We then execute a highly efficient \texttt{All-Reduce (SUM)} operation across the distributed communication group to obtain the exact global scalars $S_{\text{global}} = \mathbf{g}^\top \mathbf{m}$ and $N_{\text{global}} = \|\mathbf{m}\|^2$. 

These globally synchronized coefficients are subsequently utilized to orthogonally project the local gradient shards independently. This elegant implementation ensures that the projected distributed gradients are mathematically identical to projecting the full, unsharded parameters, demonstrating MAOP's pristine compatibility with extreme-scale FSDP infrastructures.

\subsection{Magnitude-Preserving Sparse Upcycling}
\label{app:upcycling_math}

As introduced in Sec.~\ref{sec:upcycling}, we first upcycle a dense LLM into a sparse MoE architecture. A naive weight duplication, however, would precipitate optimization instability. If all experts were exact identical copies of the dense FFN (denoted as $\hat{\mathcal{E}}$), the standard routing mechanism would double the initial activation magnitude, since the Top-2 routing weights sum to 1:
\begin{equation}
    \hat{\mathcal{E}}_{shared}(\mathbf{x}) + \sum_{i \in \text{Top-2}} g_{text, i} \, \hat{\mathcal{E}}_{text, i}(\mathbf{x}) \approx \text{FFN}_{dense}(\mathbf{x}) + \text{FFN}_{dense}(\mathbf{x}) = 2\text{FFN}_{dense}(\mathbf{x}).
\end{equation}
To strictly preserve the pre-trained dense capabilities and ensure a seamless zero-shot transition, we perform a surgical intervention \textit{exclusively during weight initialization}. We apply a deterministic scaling factor of $0.5$ to the \texttt{down\_proj} weight matrix of all upcycled experts (\textit{i.e.}, $\mathbf{W}_{down} \leftarrow 0.5 \mathbf{W}_{down}$). Since \texttt{down\_proj} constitutes the final linear projection, this initializes our actual experts to $\mathcal{E}(\mathbf{x}) \approx 0.5 \text{FFN}_{dense}(\mathbf{x})$, elegantly avoiding the cubic decay effect that would arise from scaling all internal expert matrices. Consequently, the native forward computation inherently recovers the exact original output magnitude without requiring any runtime architectural modifications:
\begin{equation}
    \mathbf{h}' = \mathcal{E}_{shared}(\mathbf{x}) + \sum_{i \in \text{Top-2}} g_{text, i} \, \mathcal{E}_{text, i}(\mathbf{x}) \approx 0.5\text{FFN}_{dense}(\mathbf{x}) + 0.5\text{FFN}_{dense}(\mathbf{x}) = \text{FFN}_{dense}(\mathbf{x}).
\end{equation}

% Empirically, this architectural initialization yields immediate dividends. 
As depicted at iteration 0 in Fig.~\ref{fig:teaser}, all evaluated sparse frameworks (Standard MoE, MoT, and Rosetta) inherit an exceptional zero-shot MMLU~\cite{hendryckstest2021} score of 52.40 without any training. This closely mirrors the officially reported 52.81 MMLU score of the dense Qwen3-0.6B Base model~\cite{yang2025qwen3}. Such near-lossless capability inheritance confirms that our magnitude-preserving upcycling flawlessly transfers pre-established knowledge into the sparse topology. This rigorous weight scaling mathematically and empirically guarantees early-stage convergence stability, allowing the subsequent 300B-token text pretraining to proceed optimally.

% ==========================================
% App B: 系统级工程实现细节 (防杠精专区)
% ==========================================
\section{System-Level Implementation Details}
\label{app:implementation}

\subsection{Two-Stage Synergistic Protection Mechanism}
\label{app:maop_stages}
% 你的原 \subsection{Integration: Two-Stage Synergistic Protection Mechanism} 放这里（解释前 5000 步的 Warmup）

While MAOP provides a mathematically rigorous guarantee against gradient conflicts, we embed it within a holistic two-stage protection mechanism to optimize the Global Shared Expert throughout the pretraining lifecycle:
\begin{itemize}
    \item \textbf{Stage 1: Warmup Gradient Shielding.} During the initial warmup phase (\textit{e.g.}, the first 5,000 steps), we explicitly detach the backward gradient paths for non-text tokens (\textit{e.g.}, ViT and VAE tokens) routed to the Shared Expert. This absolute physical isolation allows the Global Shared Expert to establish a stable, text-centric semantic foundation without early-stage high-frequency noise.
    \item \textbf{Stage 2: Permanent MAOP Intervention.} Once the semantic foundation is stabilized, the warmup shielding is lifted, and MAOP is activated permanently. It allows gradients from all modalities to update the Global Shared Expert.
\end{itemize}

\subsection{Differential Learning Rate (DiffLR) Strategy}
\label{app:difflr}
% 你的原 \section{Implementation Details: Differential Learning Rates} 放这里，带上那个伪代码 Algorithm 1！
To further stabilize the pretraining process, we adapt the differential learning rate (DiffLR) practice from Symbiotic-MoE~\cite{liu2026symbiotic}. We implement this natively within the FSDP framework by overriding the optimizer's parameter groups, as demonstrated in Algorithm~\ref{alg:difflr}.

\begin{algorithm}[h!]
\caption{Differential Learning Rate (DiffLR) Initialization under FSDP}\label{alg:difflr}
\textbf{Input:} FSDP wrapped model $f_{\theta}$, Base LR $\eta_{base}$, Generation LR $\eta_{new}$ \\
\textbf{Output:} Dual-group Optimizer $\mathcal{O}_{dual}$
\begin{algorithmic}[1]
\State $\theta_{base} \gets \emptyset$, $\theta_{gen} \gets \emptyset$
\For{\textbf{each} name, param \textbf{in} $f_{\theta}$.named\_parameters()}
    \State clean\_name $\gets$ Strip FSDP wrapper prefixes from name
    \If{clean\_name matches VAE router \textbf{or} VAE experts}
        \State $\theta_{gen} \gets \theta_{gen} \cup \{\text{param}\}$
    \Else
        \State $\theta_{base} \gets \theta_{base} \cup \{\text{param}\}$  \Comment{Includes Global Shared Experts \& Backbone}
    \EndIf
\EndFor
\State ParamGroups $\gets$[\{'params': $\theta_{gen}$, 'lr': $\eta_{new}$\}, \{'params': $\theta_{base}$, 'lr': $\eta_{base}$\}]
\State $\mathcal{O}_{dual} \gets \text{AdamW(ParamGroups)}$ \Comment{Ensures correct global LR inheritance}
\State \textbf{return} $\mathcal{O}_{dual}$
\end{algorithmic}
\end{algorithm}

% ==========================================
% App C: 极其详尽的实验与训练配置 (大一统融合区)
% ==========================================
\section{Comprehensive Pretraining Curriculum and Configurations}
% \section{Comprehensive Pretraining Curriculum}
% \section{Comprehensive Composable Pretraining Recipe}
\label{app:curriculum}
% 【军师神级合并】：把 Recipe 和 Hyperparameters 完美融合！

\subsection{Detailed Composable Pretraining Recipe}
\label{app:recipe_stages}
% 把你原来的 3 个 Plug-in 阶段精简后放在这里，说明数据比例。
% (1) Language Foundation
% (2) Visual Understanding
% (3) Visual Generation

Rosetta is fundamentally designed as a dynamic, Lego-like framework. Rather than enforcing a rigid training curriculum, we provide a composable training recipe to empirically demonstrate how modality expansion can be achieved seamlessly. To rigorously isolate the architectural impact, all empirical evaluations in this work are confined exclusively to the pretraining phase—omitting any downstream instruction tuning (SFT) or continual training (CT)—ensuring a strictly fair structural comparison.

\subsubsection{Base Modality: Language Foundation via Sparse Upcycling}
\label{sec:upcycling}

To establish a robust semantic prior for subsequent multimodal expansion, we construct sparse language foundation via sparse upcycling~\cite{komatsuzaki2022sparse} from the dense Qwen3-0.6B Base model. Specifically, the original dense FFN weights are duplicated to initialize the experts. While standard MoE uses 13 copies (12 routed + 1 shared) and MoT uses 8 copies for its understanding stream (7 routed + 1 shared), Rosetta allocates only 4 copies (3 text-specific experts + 1 global shared expert). All models enforce a Top-2 ($K=2$) routing strategy alongside the shared expert. 

To ensure optimization stability during this dense-to-sparse transition, we apply a surgical \textit{Magnitude-Preserving Initialization} by scaling the \texttt{down\_proj} weights (detailed mathematical proofs are provided in App.~\ref{app:upcycling_math}). This design allows all architectures to achieve a near-lossless capability transfer. As shown at iteration 0 in Fig.~\ref{fig:teaser}, all sparse models start with an identical MMLU score of 52.4, which closely matches the original dense Qwen3-0.6B model score of 52.8.

Following initialization, all architectures undergo rigorous pretraining on approximately 300B text tokens for 35K steps. The models are optimized using standard autoregressive cross-entropy loss alongside an expert load balancing loss~\cite{dai2024deepseekmoe}, formulated as $\mathcal{L}_{\text{total}} = \mathcal{L}_{\text{CE}} + \lambda_{\text{aux}} \mathcal{L}_{\text{aux}}$, where $\lambda_{\text{aux}} = 0.01$. 

As illustrated in Fig.~\ref{fig:teaser}, after 35K steps of text-only training (LM), standard MoE and MoT reach slightly higher MMLU scores (54.5 and 54.0, respectively) compared to Rosetta (52.5). This performance gap fundamentally aligns with established MoE scaling principles~\cite{tong2026beyond}: expanding the total expert count while holding the active parameter budget constant intrinsically increases the model's representational capacity. Consequently, since MoE and MoT allocate significantly more total experts strictly to language modeling, they naturally memorize more text data. Crucially, the objective of this stage is not to train the language models to full saturation, but rather to safely recover the dense capabilities after upcycling. This provides an absolutely fair and stabilized foundation for our primary focus: evaluating continuous multimodal expansion.

\subsubsection{Plug-in Extension: Visual Understanding}

To equip the model with visual semantics, we ``plug in'' the Visual Understanding module (Fig.~\ref{fig:pipeline} yellow block). For Rosetta, we instantiate dedicated visual experts ($\{\mathcal{E}_{vit, i}\}_{i=1}^3$) and a Top-2 router ($\mathcal{G}_{vit}$) by directly duplicating their textual counterparts, ensuring a mature warm-start initialization. For standard MoE, it continually updates its 12 modality-agnostic routed experts and 1 shared expert. For MoT, it continually updates its pre-established understanding stream (7 routed experts + 1 shared expert).

For visual feature extraction, all architectures adopt the vision encoder from Qwen3-VL-30B-Instruct~\cite{bai2025qwen3}. The ViT backbone is kept frozen, while a trainable linear projector is introduced to align the visual dimension with the LLM semantic space. Following LLaVA~\cite{liu2023visual}, we execute a progressive two-stage training across all models. First, we freeze the entire Transformer backbone and exclusively warm up the projector for 3K steps. Second, we unfreeze the backbone and jointly optimize the projector alongside the active expert routing spaces for 20K steps, consuming $\sim$4M multimodal understanding (MMU) samples sampled at a strict ratio of MMU:LM = 0.8:0.2. During this stage, the objective remains $\mathcal{L}_{\text{total}} = \mathcal{L}_{\text{CE}} + \lambda_{\text{aux}} \mathcal{L}_{\text{aux}}$, but crucially, the cross-entropy loss mask is applied exclusively to text response tokens, treating image tokens strictly as visual conditions.

\subsubsection{Plug-in Extension: Visual Generation}

To enable image generation, we ``plug in'' the Visual Generation module (Fig.~\ref{fig:pipeline} red block). For Rosetta, we instantiate six generative experts ($\{\mathcal{E}_{vae, i}\}_{i=1}^6$) and a Top-2 router ($\mathcal{G}_{vae}$) by duplicating their pre-trained textual counterparts twice, ensuring a mature warm-start initialization. To guarantee a strictly fair capacity comparison across baselines, standard MoE simply continues optimizing its pre-existing 12 modality-agnostic experts. Conversely, MoT instantiates a physically partitioned generation stream by duplicating its understanding QKV projections and copying the first 6 routed experts (with their corresponding gating slices) plus the shared expert from its understanding FFN. This rigorously enforces an identical generative expert allocation (6 routed + 1 shared) between MoT and Rosetta.

For continuous visual tokenization, all architectures adopt the FLUX.2 VAE~\cite{flux2_2025}. Applying a $2\times 2$ patchify operation yields an effective $16\times 16$ downsampling ratio with 128-dimensional latent channels. The VAE encoder is always kept frozen. In this stage, all models are jointly trained on a massive mixture of text-to-image (T2I), MMU, and LM samples for 400K steps. The overall objective seamlessly expands to $\mathcal{L}_{\text{total}} = \mathcal{L}_{\text{CE}} + \lambda_{\text{aux}} \mathcal{L}_{\text{aux}} + \lambda_{\text{img}} \mathcal{L}_{\text{flow}}$. Specifically, we employ a Flow Matching loss (velocity prediction)~\cite{lipman2022flow, liu2022flow} for image generation: $\mathcal{L}_{\text{flow}} = \mathbb{E}_{t, \mathbf{x}_0, \mathbf{x}_1} \left[ \| v_\theta(\mathbf{x}_t, t) - (\mathbf{x}_0 - \mathbf{x}_1) \|^2 \right]$, utilizing a linear flow path with log-normal timestep sampling~\cite{esser2024scaling}, setting $\lambda_{\text{img}} = 1.0$. Crucially, for Rosetta, MAOP is activated within the Global Shared Expert during this phase. By surgically neutralizing the conflicting high-frequency diffusion gradients, Rosetta successfully prevents representation overwriting, thereby preserving foundational understanding while unlocking cross-modal synergy. Detailed hyperparameters are provided in Table~\ref{tab:hyperparams}.

\subsection{Comprehensive Hyperparameters}
\label{app:hyperparams}
% 把你那个庞大的参数配置表（Table: Pretraining Configurations across Stages）放在这里！

\begin{table}[htbp]
\centering
\caption{\textbf{Pretraining Configurations across Modality Expansions.} All baselines and Rosetta are trained under identical data sequences and hyperparameter settings.}
\label{tab:hyperparams}
\resizebox{\textwidth}{!}{%
\begin{tabular}{lccc}
\toprule
\textbf{Hyperparameters} & \textbf{LM Foundation} & \textbf{+ Visual Understanding} & \textbf{+ Visual Generation} \\
\midrule
% Total Training Steps & 35,000 & 23,000 & 300,000 \\
Total Training Steps & 35,000 & 23,000 (3K Warmup + 20K Joint) & 400,000 \\
Global Batch Size & 256 & 64 & 64 \\
Peak Learning Rate & 1e-4 & Warmup (1e-4) $\rightarrow$ Joint (2e-5)  & 1e-4 (Generative) / 1e-5 (Base) \\
Optimizer & AdamW & AdamW & AdamW with DiffLR (Ours) \\
\midrule
Total Tokens / Samples & 300B Text Tokens & 4M MMU Samples & 100M T2I Samples \\
Data Sampling Ratio & LM & MMU:LM = 0.8 : 0.2 & T2I:MMU:LM = 0.6 : 0.25 : 0.15 \\
\midrule
Trainable Parameters & Full Backbone & Projector (3K steps) $\rightarrow$ Joint (20K steps) & Full Backbone \\
Frozen Modules & N/A & ViT & ViT, VAE \\
\bottomrule
\end{tabular}%
}
\end{table}

All experiments are conducted on an industrial-scale NVIDIA H20 GPU cluster. To guarantee a rigorously fair comparison, Rosetta and all baselines are subjected to the exact same hardware allocation and batch size configurations in training. Specifically, the initial language foundation pretraining is scaled across 256 GPUs with a global batch size of 256. The subsequent multimodal expansion phases (for visual understanding and visual generation) are executed on 64 GPUs with a global batch size of 64. To ensure absolute parity, all random seeds are fixed across all baselines. For visual encoding and generation, we utilize the pre-trained Qwen3-VL ViT and FLUX.2 VAE respectively, both of which remain completely frozen during our training. For the final generative expansion, we utilize a dynamic bucketing strategy centered $\approx 256 \times 256$ resolution. More hyperparameters of the pretraining regimen are provided in Table~\ref{tab:hyperparams}.

\subsection{Detailed Architectures}
\label{app:param_parity}
% 把你那个 3.77B vs 4.48B 的极其严谨的对齐表格 (Table: Comprehensive Parameter Breakdown) 放在这里！

To ensure a rigorously fair evaluation, all evaluated architectures (Standard MoE, MoT, and Rosetta) are upcycled from the identical dense foundation (Qwen3-0.6B Base) and guarantee strict active parameter parity during forward and backward passes. As detailed in Table~\ref{tab:param_breakdown}, all frameworks activate exactly 1 shared expert and 2 routed experts per token ($K=2$), resulting in a uniform active parameter count of $\sim$0.97B per token. 

However, structurally partitioned paradigms like MoT require dual attention streams and redundant expert allocations, leading to an inflated total parameter footprint (+0.71B). In contrast, Rosetta maintains the identical parameter efficiency as Standard MoE while eliminating routing collapse.

\begin{table}[htbp]
\centering
\caption{\textbf{Detailed Architecture Parameters.} Calculations are based on Qwen3-0.6B Base hyperparameters ($L=28, d_{model}=1024, d_{ffn}=3072$, vocab=157,420). All frameworks rigorously enforce strict \textit{active parameter parity} by activating exactly 3 experts (2 routed + 1 shared) and 1 attention block per token.}
\label{tab:param_breakdown}
\resizebox{0.88\linewidth}{!}{ % 稍微调大到0.85，防止新增的字数换行
\begin{tabular}{@{}lccc@{}}
\toprule
\textbf{Component} & \textbf{Standard MoE} & \textbf{Rosetta (Ours)} & \textbf{MoT (\textit{e.g.}, Bagel)} \\
\midrule
\multicolumn{4}{@{}l}{\textit{Structural Configuration}} \\
Attention Mechanism & Unified (1 Stream) & Unified (1 Stream) & Isolated (2 Streams) \\
Routed Experts Pool & 12 (Agnostic) & 3 Text + 3 ViT + 6 VAE & 7 Und. + 6 Gen. \\
Shared Experts Pool & 1 (Agnostic) & 1 (Global) & 1 Und. + 1 Gen. \\
Active Experts / Token & 2 Routed + 1 Shared & 2 Routed + 1 Shared & 2 Routed + 1 Shared \\
\midrule
\multicolumn{4}{@{}l}{\textit{Per-Layer Parameters}} \\
Attention Block & 6.29M (1 stream) & 6.29M (1 stream) & 12.58M (2 streams) \\
FFN Experts & 122.7M (13 $\times$ 9.44M) & 122.7M (13 $\times$ 9.44M) & 141.6M (15 $\times$ 9.44M) \\
Layer Norms & 0.002M & 0.002M & 0.004M \\
\cmidrule{2-4}
\textbf{Per-Layer Total} & $\approx$ \textbf{129.0M} & $\approx$ \textbf{129.0M} & $\approx$ \textbf{154.2M} \\
\midrule
\multicolumn{4}{@{}l}{\textit{Macro Parameter Budget}} \\
28-Layer Backbone & 3,612M & 3,612M & 4,318M \\
Embedding Layer & 161M & 161M & 161M \\
\cmidrule{2-4}
\textbf{Total Parameters} & \textbf{3.77B} & \textbf{3.77B} & \textbf{4.48B} \\
\midrule
\multicolumn{4}{@{}l}{\textit{Active Parameters (Per Token)}} \\
Attention Activation & 6.29M & 6.29M & 6.29M \\
FFN Activation & 28.3M (3 $\times$ 9.44M) & 28.3M (3 $\times$ 9.44M) & 28.3M (3 $\times$ 9.44M) \\
Total Active / Layer & 34.6M & 34.6M & 34.6M \\
\cmidrule{2-4}
\textbf{Total Active Params} & \textbf{$\sim$0.97B} & \textbf{$\sim$0.97B} & \textbf{$\sim$0.97B} \\
\bottomrule
\end{tabular}
}
\vspace{-1em} 
\end{table}

\vspace{0.3cm}
\begin{figure*}[h!]
    \centering
    \includegraphics[width=\textwidth]{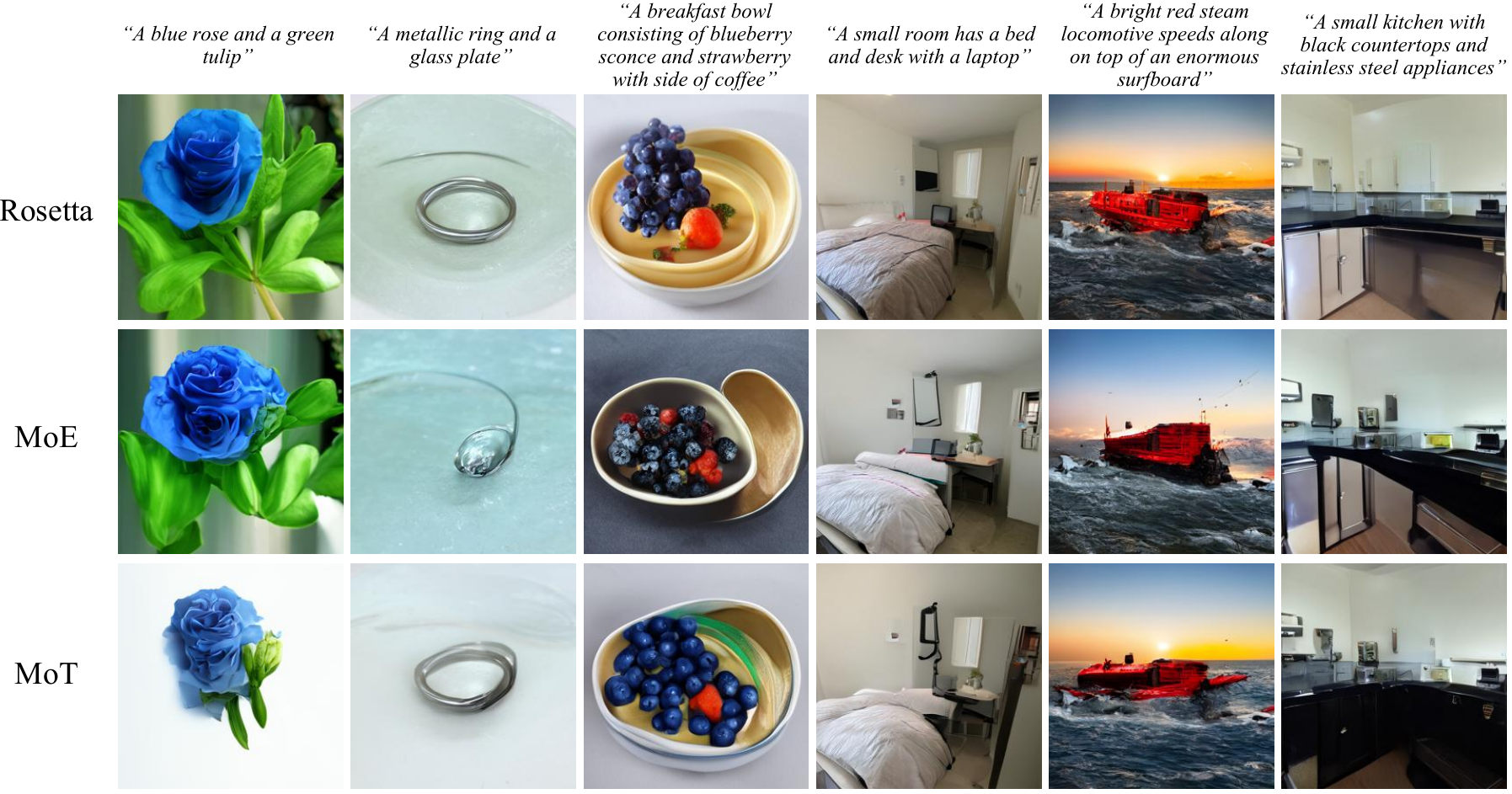}
    \includegraphics[width=\textwidth]{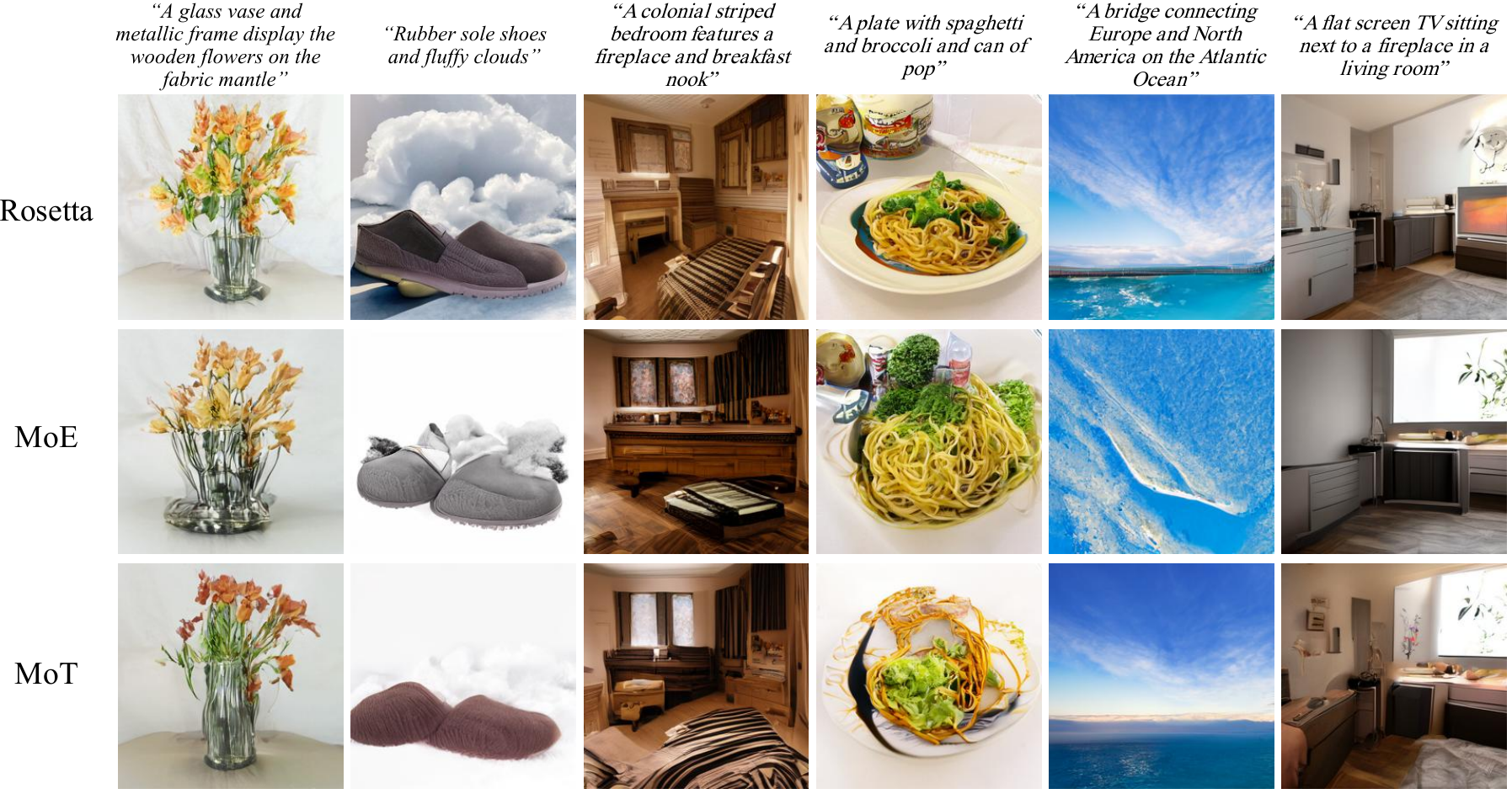}
    \caption{\textbf{Qualitative Comparisons of Image Generation.} Images generated under identical complex text prompts. \textbf{Middle Rows (MoE):} Exhibits semantic drift and visual artifacts (\textit{e.g.}, corrupted sky textures and mutated food geometries) due to representation overwriting. \textbf{Bottom Rows (MoT):} Suffers from structural collapse in indoor scenes and fails at compositional adherence (\textit{e.g.}, entirely omitting the bridge). \textbf{Top Rows (Rosetta):} Natively leverages cross-modal synergy to synthesize high-fidelity images, demonstrating precise spatial geometry, rich material textures, and strict compositional prompt adherence.}
    \label{fig:supp_images}
\end{figure*}

% ==========================================
% App D: 定海神针结果图
% ==========================================
\section{Additional Qualitative Results}
\label{app:qualitative}

To visually corroborate the catastrophic forgetting analyzed in the main text, we provide extensive qualitative comparisons in Fig.~\ref{fig:supp_images}. All models are evaluated using identical complex prompts after completing the full multimodal pretraining. 
% The generated samples unveil two drawback modes in the baseline architectures:
% \begin{itemize}[leftmargin=*]
%     \item \textbf{Semantic Drift and Artifacts (MoE):} Unconstrained routing causes generative gradients to overwrite foundational experts. Consequently, MoE frequently synthesizes severe visual artifacts (\textit{e.g.}, tearing the sky in the ``bridge'' prompt) and profound semantic mutations (\textit{e.g.}, fusing ``spaghetti and broccoli'' into an unrecognizable green mass, or melting the ``metallic ring'').
%     \item \textbf{Compositional and Structural Collapse (MoT):} Despite physical isolation, MoT's homogeneous optimization state destroys cross-modal alignment. It frequently omits key subjects entirely (\textit{e.g.}, missing the ``bridge'' or the ``green tulip'') and suffers from severe spatial geometry collapse in indoor environments (\textit{e.g.}, distorted beds and floating structures in the ``kitchen'' and ``bedroom'' prompts).
% \end{itemize}

The generated samples reveal specific challenges faced by the baseline architectures during modality expansion:
\begin{itemize}[leftmargin=*]
    \item \textbf{Semantic Drift and Visual Artifacts (MoE):} Without modality-aware constraints, continuous generative gradients can inadvertently interfere with pre-established experts. As a result, standard MoE occasionally exhibits visual artifacts (\textit{e.g.}, unnatural textures in the sky for the ``bridge'' prompt) and semantic blending (\textit{e.g.}, losing distinct object boundaries between ``spaghetti and broccoli'', or struggling with the precise geometry of the ``metallic ring'').
    
    \item \textbf{Compositional and Structural Inconsistencies (MoT):} While physical isolation protects foundational knowledge, the lack of a shared semantic bridge can limit dense cross-modal alignment. Consequently, MoT sometimes struggles with compositional completeness (\textit{e.g.}, omitting specific subjects like the ``bridge'') and may present structural distortions in complex indoor environments (\textit{e.g.}, inaccurate perspective or object placements in the ``kitchen'' and ``bedroom'' prompts).
\end{itemize}

In stark contrast, \textbf{Rosetta} neutralizes cross-modal interference. Whether rendering fine-grained material textures (\textit{e.g.}, ``metallic ring and a glass plate''), or accurate spatial geometries, Rosetta consistently synthesizes high-fidelity images with superior prompt adherence. This visually confirms the efficacy of our mechanisms in unlocking true cross-modal synergy.

\end{document}